\documentclass{tlp}
\submitted{16 October 2006} \revised{12 March 2007} \accepted{20 April 2007}

\usepackage{latexsym}
\usepackage{xspace}
\usepackage{longtable}
\def\PsfigVersion{1.9}
\ifx\undefined\psfig\else \fi

\let\LaTeXAtSign=\@
\let\@=\relax
\edef\psfigRestoreAt{\catcode`\@=\number\catcode`@\relax}
\catcode`\@=11\relax
\newwrite\@unused
\def\ps@typeout#1{{\let\protect\string\immediate\write\@unused{#1}}}
\ps@typeout{psfig/tex \PsfigVersion}

\def\figurepath{./}

\def\@nnil{\@nil}
\def\@empty{}
\def\@psdonoop#1\@@#2#3{}
\def\@psdo#1:=#2\do#3{\edef\@psdotmp{#2}\ifx\@psdotmp\@empty \else
    \expandafter\@psdoloop#2,\@nil,\@nil\@@#1{#3}\fi}
\def\@psdoloop#1,#2,#3\@@#4#5{\def#4{#1}\ifx #4\@nnil \else
       #5\def#4{#2}\ifx #4\@nnil \else#5\@ipsdoloop #3\@@#4{#5}\fi\fi}
\def\@ipsdoloop#1,#2\@@#3#4{\def#3{#1}\ifx #3\@nnil 
       \let\@nextwhile=\@psdonoop \else
      #4\relax\let\@nextwhile=\@ipsdoloop\fi\@nextwhile#2\@@#3{#4}}
\def\@tpsdo#1:=#2\do#3{\xdef\@psdotmp{#2}\ifx\@psdotmp\@empty \else
    \@tpsdoloop#2\@nil\@nil\@@#1{#3}\fi}
\def\@tpsdoloop#1#2\@@#3#4{\def#3{#1}\ifx #3\@nnil 
       \let\@nextwhile=\@psdonoop \else
      #4\relax\let\@nextwhile=\@tpsdoloop\fi\@nextwhile#2\@@#3{#4}}
\ifx\undefined\fbox
\newdimen\fboxrule
\newdimen\fboxsep
\newdimen\ps@tempdima
\newbox\ps@tempboxa
\long\def\fbox#1{\leavevmode\setbox\ps@tempboxa\hbox{#1}\ps@tempdima\fboxrule
    \advance\ps@tempdima \fboxsep \advance\ps@tempdima \dp\ps@tempboxa
   \hbox{\lower \ps@tempdima\hbox
  {\vbox{\hrule height \fboxrule
          \hbox{\vrule width \fboxrule \hskip\fboxsep
          \vbox{\vskip\fboxsep \box\ps@tempboxa\vskip\fboxsep}\hskip 
                 \fboxsep\vrule width \fboxrule}
                 \hrule height \fboxrule}}}}
\fi
\newread\ps@stream
\newif\ifnot@eof       
\newif\if@noisy        
\newif\if@atend        
\newif\if@psfile       
{\catcode`\%=12\global\gdef\epsf@start{
\def\epsf@PS{PS}
\def\epsf@getbb#1{%
\openin\ps@stream=#1
\ifeof\ps@stream\ps@typeout{Error, File #1 not found}\else
   {\not@eoftrue \chardef\other=12
    \def\do##1{\catcode`##1=\other}\dospecials \catcode`\ =10
    \loop
       \if@psfile
	  \read\ps@stream to \epsf@fileline
       \else{
	  \obeyspaces
          \read\ps@stream to \epsf@tmp\global\let\epsf@fileline\epsf@tmp}
       \fi
       \ifeof\ps@stream\not@eoffalse\else
       \if@psfile\else
       \expandafter\epsf@test\epsf@fileline:. \\%
       \fi
          \expandafter\epsf@aux\epsf@fileline:. \\%
       \fi
   \ifnot@eof\repeat
   }\closein\ps@stream\fi}%
\long\def\epsf@test#1#2#3:#4\\{\def\epsf@testit{#1#2}
			\ifx\epsf@testit\epsf@start\else
\ps@typeout{Warning! File does not start with `\epsf@start'.  It may not be a PostScript file.}
			\fi
			\@psfiletrue} 
{\catcode`\%=12\global\let\epsf@percent=
\long\def\epsf@aux#1#2:#3\\{\ifx#1\epsf@percent
   \def\epsf@testit{#2}\ifx\epsf@testit\epsf@bblit
	\@atendfalse
        \epsf@atend #3 . \\%
	\if@atend	
	   \if@verbose{
		\ps@typeout{psfig: found `(atend)'; continuing search}
	   }\fi
        \else
        \epsf@grab #3 . . . \\%
        \not@eoffalse
        \global\no@bbfalse
        \fi
   \fi\fi}%
\def\epsf@grab #1 #2 #3 #4 #5\\{%
   \global\def\epsf@llx{#1}\ifx\epsf@llx\empty
      \epsf@grab #2 #3 #4 #5 .\\\else
   \global\def\epsf@lly{#2}%
   \global\def\epsf@urx{#3}\global\def\epsf@ury{#4}\fi}%
\def\epsf@atendlit{(atend)} 
\def\epsf@atend #1 #2 #3\\{%
   \def\epsf@tmp{#1}\ifx\epsf@tmp\empty
      \epsf@atend #2 #3 .\\\else
   \ifx\epsf@tmp\epsf@atendlit\@atendtrue\fi\fi}

\chardef\psletter = 11 
\chardef\other = 12

\newif \ifdebug 
\newif\ifc@mpute 
\c@mputetrue 

\let\then = \relax
\def\r@dian{pt }
\let\r@dians = \r@dian
\let\dimensionless@nit = \r@dian
\let\dimensionless@nits = \dimensionless@nit
\def\internal@nit{sp }
\let\internal@nits = \internal@nit
\newif\ifstillc@nverging
\def \Mess@ge #1{\ifdebug \then \message {#1} \fi}

{ 
	\catcode `\@ = \psletter
	\gdef \nodimen {\expandafter \n@dimen \the \dimen}
	\gdef \term #1 #2 #3%
	       {\edef \t@ {\the #1}
		\edef \t@@ {\expandafter \n@dimen \the #2\r@dian}%
		\t@rm {\t@} {\t@@} {#3}%
	       }
	\gdef \t@rm #1 #2 #3%
	       {{%
		\count 0 = 0
		\dimen 0 = 1 \dimensionless@nit
		\dimen 2 = #2\relax
		\Mess@ge {Calculating term #1 of \nodimen 2}%
		\loop
		\ifnum	\count 0 < #1
		\then	\advance \count 0 by 1
			\Mess@ge {Iteration \the \count 0 \space}%
			\Multiply \dimen 0 by {\dimen 2}%
			\Mess@ge {After multiplication, term = \nodimen 0}%
			\Divide \dimen 0 by {\count 0}%
			\Mess@ge {After division, term = \nodimen 0}%
		\repeat
		\Mess@ge {Final value for term #1 of 
				\nodimen 2 \space is \nodimen 0}%
		\xdef \Term {#3 = \nodimen 0 \r@dians}%
		\aftergroup \Term
	       }}
	\catcode `\p = \other
	\catcode `\t = \other
	\gdef \n@dimen #1pt{#1} 
}

\def \Divide #1by #2{\divide #1 by #2} 

\def \Multiply #1by #2
       {{
	\count 0 = #1\relax
	\count 2 = #2\relax
	\count 4 = 65536
	\Mess@ge {Before scaling, count 0 = \the \count 0 \space and
			count 2 = \the \count 2}%
	\ifnum	\count 0 > 32767 
	\then	\divide \count 0 by 4
		\divide \count 4 by 4
	\else	\ifnum	\count 0 < -32767
		\then	\divide \count 0 by 4
			\divide \count 4 by 4
		\else
		\fi
	\fi
	\ifnum	\count 2 > 32767 
	\then	\divide \count 2 by 4
		\divide \count 4 by 4
	\else	\ifnum	\count 2 < -32767
		\then	\divide \count 2 by 4
			\divide \count 4 by 4
		\else
		\fi
	\fi
	\multiply \count 0 by \count 2
	\divide \count 0 by \count 4
	\xdef \product {#1 = \the \count 0 \internal@nits}%
	\aftergroup \product
       }}

\def\r@duce{\ifdim\dimen0 > 90\r@dian \then   
		\multiply\dimen0 by -1
		\advance\dimen0 by 180\r@dian
		\r@duce
	    \else \ifdim\dimen0 < -90\r@dian \then  
		\advance\dimen0 by 360\r@dian
		\r@duce
		\fi
	    \fi}

\def\Sine#1%
       {{%
	\dimen 0 = #1 \r@dian
	\r@duce
	\ifdim\dimen0 = -90\r@dian \then
	   \dimen4 = -1\r@dian
	   \c@mputefalse
	\fi
	\ifdim\dimen0 = 90\r@dian \then
	   \dimen4 = 1\r@dian
	   \c@mputefalse
	\fi
	\ifdim\dimen0 = 0\r@dian \then
	   \dimen4 = 0\r@dian
	   \c@mputefalse
	\fi
	\ifc@mpute \then
		\divide\dimen0 by 180
		\dimen0=3.141592654\dimen0
		\dimen 2 = 3.1415926535897963\r@dian 
		\divide\dimen 2 by 2 
		\Mess@ge {Sin: calculating Sin of \nodimen 0}%
		\count 0 = 1 
		\dimen 2 = 1 \r@dian 
		\dimen 4 = 0 \r@dian 
		\loop
			\ifnum	\dimen 2 = 0 
			\then	\stillc@nvergingfalse 
			\else	\stillc@nvergingtrue
			\fi
			\ifstillc@nverging 
			\then	\term {\count 0} {\dimen 0} {\dimen 2}%
				\advance \count 0 by 2
				\count 2 = \count 0
				\divide \count 2 by 2
				\ifodd	\count 2 
				\then	\advance \dimen 4 by \dimen 2
				\else	\advance \dimen 4 by -\dimen 2
				\fi
		\repeat
	\fi		
			\xdef \sine {\nodimen 4}%
       }}

\def\Cosine#1{\ifx\sine\UnDefined\edef\Savesine{\relax}\else
		             \edef\Savesine{\sine}\fi
	{\dimen0=#1\r@dian\advance\dimen0 by 90\r@dian
	 \Sine{\nodimen 0}
	 \xdef\cosine{\sine}
	 \xdef\sine{\Savesine}}}	      

\def\psdraft{
	\def\@psdraft{0}
}
\def\psfull{
	\def\@psdraft{100}
}

\psfull

\newif\if@scalefirst
\def\psscalefirst{\@scalefirsttrue}
\def\psrotatefirst{\@scalefirstfalse}
\psrotatefirst

\newif\if@draftbox
\def\psnodraftbox{
	\@draftboxfalse
}
\def\psdraftbox{
	\@draftboxtrue
}
\@draftboxtrue

\newif\if@prologfile
\newif\if@postlogfile
\def\pssilent{
	\@noisyfalse
}
\def\psnoisy{
	\@noisytrue
}
\psnoisy
\newif\if@bbllx
\newif\if@bblly
\newif\if@bburx
\newif\if@bbury
\newif\if@height
\newif\if@width
\newif\if@rheight
\newif\if@rwidth
\newif\if@angle
\newif\if@clip
\newif\if@verbose
\def\@p@@sclip#1{\@cliptrue}

\newif\if@decmpr

\def\@p@@sfigure#1{\def\@p@sfile{null}\def\@p@sbbfile{null}
	        \openin1=#1.bb
		\ifeof1\closein1
	        	\openin1=\figurepath#1.bb
			\ifeof1\closein1
			        \openin1=#1
				\ifeof1\closein1%
				       \openin1=\figurepath#1
					\ifeof1
					   \ps@typeout{Error, File #1 not found}
						\if@bbllx\if@bblly
				   		\if@bburx\if@bbury
			      				\def\@p@sfile{#1}%
			      				\def\@p@sbbfile{#1}%
							\@decmprfalse
				  	   	\fi\fi\fi\fi
					\else\closein1
				    		\def\@p@sfile{\figurepath#1}%
				    		\def\@p@sbbfile{\figurepath#1}%
						\@decmprfalse
	                       		\fi%
			 	\else\closein1%
					\def\@p@sfile{#1}
					\def\@p@sbbfile{#1}
					\@decmprfalse
			 	\fi
			\else
				\def\@p@sfile{\figurepath#1}
				\def\@p@sbbfile{\figurepath#1.bb}
				\@decmprtrue
			\fi
		\else
			\def\@p@sfile{#1}
			\def\@p@sbbfile{#1.bb}
			\@decmprtrue
		\fi}

\def\@p@@sfile#1{\@p@@sfigure{#1}}

\def\@p@@sbbllx#1{
		\@bbllxtrue
		\dimen100=#1
		\edef\@p@sbbllx{\number\dimen100}
}
\def\@p@@sbblly#1{
		\@bbllytrue
		\dimen100=#1
		\edef\@p@sbblly{\number\dimen100}
}
\def\@p@@sbburx#1{
		\@bburxtrue
		\dimen100=#1
		\edef\@p@sbburx{\number\dimen100}
}
\def\@p@@sbbury#1{
		\@bburytrue
		\dimen100=#1
		\edef\@p@sbbury{\number\dimen100}
}
\def\@p@@sheight#1{
		\@heighttrue
		\dimen100=#1
   		\edef\@p@sheight{\number\dimen100}
}
\def\@p@@swidth#1{
		\@widthtrue
		\dimen100=#1
		\edef\@p@swidth{\number\dimen100}
}
\def\@p@@srheight#1{
		\@rheighttrue
		\dimen100=#1
		\edef\@p@srheight{\number\dimen100}
}
\def\@p@@srwidth#1{
		\@rwidthtrue
		\dimen100=#1
		\edef\@p@srwidth{\number\dimen100}
}
\def\@p@@sangle#1{
		\@angletrue
		\edef\@p@sangle{#1} 
}
\def\@p@@ssilent#1{ 
		\@verbosefalse
}
\def\@p@@sprolog#1{\@prologfiletrue\def\@prologfileval{#1}}
\def\@p@@spostlog#1{\@postlogfiletrue\def\@postlogfileval{#1}}
\def\@cs@name#1{\csname #1\endcsname}
\def\@setparms#1=#2,{\@cs@name{@p@@s#1}{#2}}
\def\ps@init@parms{
		\@bbllxfalse \@bbllyfalse
		\@bburxfalse \@bburyfalse
		\@heightfalse \@widthfalse
		\@rheightfalse \@rwidthfalse
		\def\@p@sbbllx{}\def\@p@sbblly{}
		\def\@p@sbburx{}\def\@p@sbbury{}
		\def\@p@sheight{}\def\@p@swidth{}
		\def\@p@srheight{}\def\@p@srwidth{}
		\def\@p@sangle{0}
		\def\@p@sfile{} \def\@p@sbbfile{}
		\def\@p@scost{10}
		\def\@sc{}
		\@prologfilefalse
		\@postlogfilefalse
		\@clipfalse
		\if@noisy
			\@verbosetrue
		\else
			\@verbosefalse
		\fi
}
\def\parse@ps@parms#1{
	 	\@psdo\@psfiga:=#1\do
		   {\expandafter\@setparms\@psfiga,}}
\newif\ifno@bb
\def\bb@missing{
	\if@verbose{
		\ps@typeout{psfig: searching \@p@sbbfile \space  for bounding box}
	}\fi
	\no@bbtrue
	\epsf@getbb{\@p@sbbfile}
        \ifno@bb \else \bb@cull\epsf@llx\epsf@lly\epsf@urx\epsf@ury\fi
}	
\def\bb@cull#1#2#3#4{
	\dimen100=#1 bp\edef\@p@sbbllx{\number\dimen100}
	\dimen100=#2 bp\edef\@p@sbblly{\number\dimen100}
	\dimen100=#3 bp\edef\@p@sbburx{\number\dimen100}
	\dimen100=#4 bp\edef\@p@sbbury{\number\dimen100}
	\no@bbfalse
}
\newdimen\p@intvaluex
\newdimen\p@intvaluey
\def\rotate@#1#2{{\dimen0=#1 sp\dimen1=#2 sp
		  \global\p@intvaluex=\cosine\dimen0
		  \dimen3=\sine\dimen1
		  \global\advance\p@intvaluex by -\dimen3
		  \global\p@intvaluey=\sine\dimen0
		  \dimen3=\cosine\dimen1
		  \global\advance\p@intvaluey by \dimen3
		  }}
\def\compute@bb{
		\no@bbfalse
		\if@bbllx \else \no@bbtrue \fi
		\if@bblly \else \no@bbtrue \fi
		\if@bburx \else \no@bbtrue \fi
		\if@bbury \else \no@bbtrue \fi
		\ifno@bb \bb@missing \fi
		\ifno@bb \ps@typeout{FATAL ERROR: no bb supplied or found}
			\no-bb-error
		\fi
		\count203=\@p@sbburx
		\count204=\@p@sbbury
		\advance\count203 by -\@p@sbbllx
		\advance\count204 by -\@p@sbblly
		\edef\ps@bbw{\number\count203}
		\edef\ps@bbh{\number\count204}
		\if@angle 
			\Sine{\@p@sangle}\Cosine{\@p@sangle}
	        	{\dimen100=\maxdimen\xdef\r@p@sbbllx{\number\dimen100}
					    \xdef\r@p@sbblly{\number\dimen100}
			                    \xdef\r@p@sbburx{-\number\dimen100}
					    \xdef\r@p@sbbury{-\number\dimen100}}
                        \def\minmaxtest{
			   \ifnum\number\p@intvaluex<\r@p@sbbllx
			      \xdef\r@p@sbbllx{\number\p@intvaluex}\fi
			   \ifnum\number\p@intvaluex>\r@p@sbburx
			      \xdef\r@p@sbburx{\number\p@intvaluex}\fi
			   \ifnum\number\p@intvaluey<\r@p@sbblly
			      \xdef\r@p@sbblly{\number\p@intvaluey}\fi
			   \ifnum\number\p@intvaluey>\r@p@sbbury
			      \xdef\r@p@sbbury{\number\p@intvaluey}\fi
			   }
			\rotate@{\@p@sbbllx}{\@p@sbblly}
			\minmaxtest
			\rotate@{\@p@sbbllx}{\@p@sbbury}
			\minmaxtest
			\rotate@{\@p@sbburx}{\@p@sbblly}
			\minmaxtest
			\rotate@{\@p@sbburx}{\@p@sbbury}
			\minmaxtest
			\edef\@p@sbbllx{\r@p@sbbllx}\edef\@p@sbblly{\r@p@sbblly}
			\edef\@p@sbburx{\r@p@sbburx}\edef\@p@sbbury{\r@p@sbbury}
		\fi
		\count203=\@p@sbburx
		\count204=\@p@sbbury
		\advance\count203 by -\@p@sbbllx
		\advance\count204 by -\@p@sbblly
		\edef\@bbw{\number\count203}
		\edef\@bbh{\number\count204}
}
\def\in@hundreds#1#2#3{\count240=#2 \count241=#3
		     \count100=\count240	
		     \divide\count100 by \count241
		     \count101=\count100
		     \multiply\count101 by \count241
		     \advance\count240 by -\count101
		     \multiply\count240 by 10
		     \count101=\count240	
		     \divide\count101 by \count241
		     \count102=\count101
		     \multiply\count102 by \count241
		     \advance\count240 by -\count102
		     \multiply\count240 by 10
		     \count102=\count240	
		     \divide\count102 by \count241
		     \count200=#1\count205=0
		     \count201=\count200
			\multiply\count201 by \count100
		 	\advance\count205 by \count201
		     \count201=\count200
			\divide\count201 by 10
			\multiply\count201 by \count101
			\advance\count205 by \count201
		     \count201=\count200
			\divide\count201 by 100
			\multiply\count201 by \count102
			\advance\count205 by \count201
		     \edef\@result{\number\count205}
}
\def\compute@wfromh{
		\in@hundreds{\@p@sheight}{\@bbw}{\@bbh}
		\edef\@p@swidth{\@result}
}
\def\compute@hfromw{
	        \in@hundreds{\@p@swidth}{\@bbh}{\@bbw}
		\edef\@p@sheight{\@result}
}
\def\compute@handw{
		\if@height 
			\if@width
			\else
				\compute@wfromh
			\fi
		\else 
			\if@width
				\compute@hfromw
			\else
				\edef\@p@sheight{\@bbh}
				\edef\@p@swidth{\@bbw}
			\fi
		\fi
}
\def\compute@resv{
		\if@rheight \else \edef\@p@srheight{\@p@sheight} \fi
		\if@rwidth \else \edef\@p@srwidth{\@p@swidth} \fi
}
\def\compute@sizes{
	\compute@bb
	\if@scalefirst\if@angle
	\if@width
	   \in@hundreds{\@p@swidth}{\@bbw}{\ps@bbw}
	   \edef\@p@swidth{\@result}
	\fi
	\if@height
	   \in@hundreds{\@p@sheight}{\@bbh}{\ps@bbh}
	   \edef\@p@sheight{\@result}
	\fi
	\fi\fi
	\compute@handw
	\compute@resv}

\def\psfig#1{\vbox {
	\ps@init@parms
	\parse@ps@parms{#1}
	\compute@sizes
	\ifnum\@p@scost<\@psdraft{
		\special{ps::[begin] 	\@p@swidth \space \@p@sheight \space
				\@p@sbbllx \space \@p@sbblly \space
				\@p@sbburx \space \@p@sbbury \space
				startTexFig \space }
		\if@angle
			\special {ps:: \@p@sangle \space rotate \space} 
		\fi
		\if@clip{
			\if@verbose{
				\ps@typeout{(clip)}
			}\fi
			\special{ps:: doclip \space }
		}\fi
		\if@prologfile
		    \special{ps: plotfile \@prologfileval \space } \fi
		\if@decmpr{
			\if@verbose{
				\ps@typeout{psfig: including \@p@sfile.Z \space }
			}\fi
			\special{ps: plotfile "`zcat \@p@sfile.Z" \space }
		}\else{
			\if@verbose{
				\ps@typeout{psfig: including \@p@sfile \space }
			}\fi
			\special{ps: plotfile \@p@sfile \space }
		}\fi
		\if@postlogfile
		    \special{ps: plotfile \@postlogfileval \space } \fi
		\special{ps::[end] endTexFig \space }
		\vbox to \@p@srheight sp{
			\hbox to \@p@srwidth sp{
				\hss
			}
		\vss
		}
	}\else{
		\if@draftbox{		
			\hbox{\frame{\vbox to \@p@srheight sp{
			\vss
			\hbox to \@p@srwidth sp{ \hss \@p@sfile \hss }
			\vss
			}}}
		}\else{
			\vbox to \@p@srheight sp{
			\vss
			\hbox to \@p@srwidth sp{\hss}
			\vss
			}
		}\fi

	}\fi
}}
\psfigRestoreAt
\let\@=\LaTeXAtSign

 \usepackage{epic}
 \usepackage{eepic}
 \usepackage{epsfig}
 \usepackage{url}
 \usepackage{xspace}
 \usepackage{latexsym}

\newif\ifreport

\newcommand{\TODO}[1]{{\large\bf TODO: }#1\ensuremath{\Box}}
\newcommand{\nop}[1]{}

\newtheorem{theorem}{Theorem}[section]
\newtheorem{example}[theorem]{Example}

\newtheorem{definition}[theorem]{Definition}

\newcommand{\etal}{\textit{et al.}\xspace}

\newcommand{\tabstretch}{\raisebox{0ex}[3ex][2ex]{}}

\newcommand{\mathem}[1]{\emph{\ensuremath{#1}}}

\newcommand{\set}[1]{\{ #1 \}}

\newcommand{\Pol}{{\rm P}}
\newcommand{\NP}{\textrm{NP}\xspace}
\newcommand{\CONP}{\textrm{co-NP}\xspace}
\newcommand{\SigmaP}[1]{\ensuremath{\Sigma_{#1}^P}}
\newcommand{\PiP}[1]{{\Pi}_{#1}^{P}}
\newcommand{\DeltaP}[1]{\ensuremath{\Delta_{#1}^P}}

\newcommand{\Or}{\ensuremath{\mathtt{\,v\,}}\xspace}
\newcommand{\derives}{\mbox{\,\texttt{:\hspace{-0.15em}-}}\,\xspace}
\newcommand{\wderives}{\ensuremath{:\sim}}
\newcommand{\uneq}{\ensuremath{<>}}
\newcommand{\Comma}{\ \ ;\ \ }

\newcommand{\p}{\ensuremath{{\cal P}}}
\newcommand{\n}{\ensuremath{{\cal N}}}
\newcommand{\GP}{\ensuremath{Ground(\p)}}
\newcommand{\GRules}{\ensuremath{GroundRules(\p)}}
\newcommand{\GWC}{\ensuremath{GroundWC(\p)}}
\newcommand{\BP}{\ensuremath{B_{\p}}}
\newcommand{\UP}{\ensuremath{U_{\p}}}

\newcommand{\R}{\ensuremath{r}}
\newcommand{\GR}{\ensuremath{Ground(\R)}}
\newcommand{\Gaweak}{\ensuremath{Ground(\aweak)}}
\newcommand{\HR}{\ensuremath{H(\R)}}
\newcommand{\BR}{\ensuremath{B(\R)}}
\newcommand{\BpR}{\ensuremath{B^+(\R)}}
\newcommand{\BnR}{\ensuremath{B^-(\R)}}

\newcommand{\tneg}{\ensuremath{\neg}}
\newcommand{\naf}{\ensuremath{\mathrm{not}}\xspace}

\newcommand{\dlv}{{\small DLV}\xspace}
\newcommand{\dlvdb}{{\small DLV}$^{DB}$\xspace}
\newcommand{\dlvio}{{\small DLV}$^{IO}$\xspace}
\newcommand{\dlvf}[1]{\mbox{{\sc DLV}{\rm[}$#1${\rm]}}\xspace}
\newcommand{\K}{\ensuremath{\cal K}\xspace}
\newcommand{\dlvk}{\texttt{\small{DLV}$^{\K}$}\xspace}
\newcommand{\Kc}{\ensuremath{\mathcal{K}^c}\xspace}
\newcommand{\gnt}{GnT\xspace}
\newcommand{\NoMoRe}{NoMoRe\xspace}

\newcommand{\gco}{{\bf \small{GCO}}\xspace}

\newcommand{\DLPI}{\ensuremath{\mathrm{DLP}^<}}
\newcommand{\DLPF}[1]{\mbox{{\sc DLP}{\rm[}$#1${\rm]}}\xspace}

\newcommand{\tuple}[1]{\langle#1\rangle}
\newcommand{\Q}[1]{$\cal Q$}

\def\WP{{\cal W}_{\p}}
\def\WPinf{{\cal W}_{\p}^{\omega}(\emptyset)}

\newcommand{\nafsymbol}{\ensuremath{\mathtt{not}}}
\newcommand{\intnum}{\ensuremath{I}}
\newcommand{\rational}{\ensuremath{Q}}

\newcommand{\varset}{\ensuremath{\sigma_{var}}\xspace}
\newcommand{\constset}{\ensuremath{\sigma_{const}}\xspace}
\newcommand{\predset}{\ensuremath{\sigma_{pred}}\xspace}
\newcommand{\termset}{\ensuremath{\sigma_{term}}\xspace}

\newcommand{\arity}[1]{\ensuremath{arity(#1)}}

\newcommand{\q}{\ensuremath{{\cal Q}}}
\newcommand{\guessprog}{\ensuremath{{\cal G}}}
\newcommand{\checkprog}{\ensuremath{{\cal C}}}
\newcommand{\optprog}{\ensuremath{{\cal O}}}
\newcommand{\factprog}{\ensuremath{{\cal F}}}

\newcommand{\base}[1]{\ensuremath{B_{#1}}}
\newcommand{\universe}[1]{\ensuremath{U_{#1}}}

\newcommand{\ground}[1]{\ensuremath{Ground(#1)}}

\newcommand{\programs}[1]{\ensuremath{\Pi_{#1}}}

\newcommand{\arule}[1][]{\ensuremath{r_{#1}}}
\newcommand{\aconstr}[1][]{\ensuremath{c_{#1}}}
\newcommand{\aquery}[1][]{\ensuremath{q_{#1}}}
\newcommand{\aprog}[1][]{\ensuremath{\p_{#1}}}
\newcommand{\anint}[1][]{\ensuremath{I_{#1}}}
\newcommand{\apint}[1][]{\ensuremath{{\cal I}_{#1}}}
\newcommand{\aweak}[1][]{\ensuremath{w_{#1}}}
\newcommand{\apjnt}[1][]{\ensuremath{{\cal J}_{#1}}}
\newcommand{\incint}{\ensuremath{\Box}}

\newcommand{\bravecons}{\ensuremath{\models_b}}
\newcommand{\cautiouscons}{\ensuremath{\models_c}}

\newcommand{\head}[1]{\ensuremath{H(#1)}}
\newcommand{\body}[1]{\ensuremath{B(#1)}}
\newcommand{\posbody}[1]{\ensuremath{B^+(#1)}}
\newcommand{\negbody}[1]{\ensuremath{B^-(#1)}}
\newcommand{\litbody}[1]{\ensuremath{{\cal B}(#1)}}

\newcommand{\AS}[1]{\ensuremath{{\cal AS}(#1)}}
\newcommand{\OAS}[1]{\ensuremath{{\cal OAS}(#1)}}
\newcommand{\ASI}[2]{\ensuremath{{\cal AS}(#1,#2)}}

\newcommand{\wmax}[1]{\ensuremath{w_{max}^{#1}}}
\newcommand{\lmax}[1]{\ensuremath{l_{max}^{#1}}}

\newcommand{\val}[2]{\ensuremath{val_{#1}(#2)}}
\newcommand{\valh}[2]{\ensuremath{val^H_{#1}(#2)}}
\newcommand{\valb}[2]{\ensuremath{val^B_{#1}(#2)}}

\newcommand{\WC}[1]{\ensuremath{WC(#1)}}
\newcommand{\Rules}[1]{\ensuremath{Rules(#1)}}

\newcommand{\rulesetsep}{\ \ ;\ \ }

\newcommand{\dlfact}[1]{\ensuremath{#1.}}
\newcommand{\dlquery}[1]{\ensuremath{#1?}}
\newcommand{\dlrule}[2]{\ensuremath{#1 \derives #2.}}
\newcommand{\dlconstraint}[1]{\ensuremath{\derives #1.}}
\newcommand{\dlweakconstraint}[3]{\ensuremath{\wderives #1.\ [#2:#3]}}

\newcommand{\dlstrictfact}[1]{\ensuremath{#1!}}
\newcommand{\dlstrictrule}[2]{\ensuremath{#1 \derives #2!}}

\newcommand{\dlalignedfact}[1]{\code{#1.}}
\newcommand{\dlalignedquery}[1]{\code{#1?}}
\newcommand{\dlalignrule}[2]{\code{#1 \derives } \=\code{#2.}}
\newcommand{\dlalignedrule}[2]{\code{#1 \derives } \>\code{#2.}}
\newcommand{\dlalignedincompleterule}[2]{\code{#1 \derives } \=\code{#2}}
\newcommand{\dlalignedstubrule}[1]{\> \code{#1.}}
\newcommand{\dlalignconstraint}[1]{\code{\derives } \=\code{#1.}}
\newcommand{\dlalignedconstraint}[1]{\code{\derives } \>\code{#1.}}
\newcommand{\dlalignedincompleteconstraint}[1]{\code{\derives } \=\code{#1}}
\newcommand{\dlalignedstubconstraint}[1]{\>\code{#1.}}
\newcommand{\dlalignweakconstraint}[2]{\code{\wderives} \=\code{#1. [#2]}}
\newcommand{\dlalignedweakconstraint}[2]{\code{\wderives} \>\code{#1. [#2]}}

\newcommand{\displacedcode}[1]{\begin{center}\ensuremath{\mathtt{#1}}\end{center}}

\newcommand{\inlinekrule}[1]{\texttt{#1}}

\newenvironment{simpleprogram}[1][]
   {\vspace{-0.5ex}\begin{itemize}\item[]
      \tt
      \begin{tabbing}
      \code{#1}\ \= \kill
   }
   {\end{tabbing}\end{itemize}\vspace{-3ex}}

\newcommand{\spi}[2][]{\code{#1}\>\code{#2}\\}
\newcommand{\spfsi}[2][]{\footnotesize\code{#1}\>\footnotesize\code{#2}\\[-0.8ex]}

\newenvironment{simplealignedprogramstub}[1][]
   {\vspace{-0ex}
      \begin{tabbing}
      #1\kill
   }
   {\end{tabbing}
\vspace{-6ex}}

 \newcommand{\sapi}[1]{#1\\}

\newenvironment{sublabeledprogram}[1][]
   {\begin{array}{ll}\setlength{\arraycolsep}{0pt}}
   {\end{array}}

\newcommand{\lpi}[2]{\left.\begin{minipage}{\textwidth}\begin{simpleprogram}#1\end{simpleprogram}\vspace{-3ex}\end{minipage}\right\} \mathbf{#2}\\[2ex]}
\newcommand{\lpsi}[1]{\spi{#1}}

\newcommand{\gt}{\ensuremath{>}}
\newcommand{\lt}{\ensuremath{<}}

\newcommand{\code}[1]{\ensuremath{#1}}
\newenvironment{dlvcode}
  {\begin{displaymath}\begin{array}{l}}
  {\end{array}\end{displaymath}}

\newcommand{\emptycodeline}{\\[-3mm]}

\newcommand{\countagg}{\ensuremath{\mathtt{\# count}}}
\newcommand{\sumagg}{\ensuremath{\mathtt{\# sum}}}
\newcommand{\minagg}{\ensuremath{\mathtt{\# min}}}
\newcommand{\maxagg}{\ensuremath{\mathtt{\# max}}}
\newcommand{\timesagg}{\ensuremath{\mathtt{\# times}}}
\newcommand{\anyagg}{\ensuremath{\mathtt{\# any}}}
\newcommand{\avgagg}{\ensuremath{\mathtt{\# avg}}}

\def\F{$\mathcal{F}$}
\def\P{\ensuremath{\mathcal{P}}}
\def\S{\ensuremath{\mathcal{S}}}
\def\A{$A$}
\newcommand{\dlva}{{\sc DLV}\ensuremath{^\mathcal{A}}\xspace}
\newcommand{\DLP}{{\sc DLP}\xspace}
\newcommand{\DLPA}{{\sc DLP}\ensuremath{^\mathcal{A}}\xspace}
\newcommand{\edb}{{\sc EDB(\p)}\xspace}
\newcommand{\idb}{{\sc IDB(\p)}\xspace}

\newcommand{\vpiccolo}{\vspace*{-0.2cm}}
\newcommand{\vmedio}{\vspace*{-0.4cm}}
\newcommand{\vgrande}{\vspace*{-0.6cm}}

\newcommand{\cmpitem}{\ensuremath{\bullet}\xspace}
\newcommand{\cmpenum}[2]{\ensuremath{(#1_{#2})}}

\newcommand{\UPn}{\ensuremath{U_{\p}^{\n}}\xspace}
\newcommand{\UPs}{\ensuremath{U_{\p}^{\s}}\xspace}

\newcommand{\multi}[1]{\ensuremath{\overline{2}^{#1}}}

\newcommand{\qed}{\hfill $\Box$}

\newcommand{\MM}{{\cal MM}}
\newcommand{\SM}{{\cal SM}}

\newcommand{\kw}[1]{{\bf #1}}
\def\kif{\kw{if}}
\def\kthen{\kw{then}}
\def\kelse{\kw{else}}
\def\kfor{\kw{for}}
\def\kwhile{\kw{while}}
\def\kdo{\kw{do}}
\def\kbegin{\kw{begin}}
\def\kend{\kw{end}}
\def\kelseif{\kw{else} \kw{if}}
\def\kto{\kw{to}}
\def\krepeat{\kw{repeat}}
\def\kuntil{\kw{until}}
\def\kreturn{\kw{return}}

\newcommand{\conj}{\mathit{conj}}
\newcommand{\const}[1]{\textsf{#1}}
\newcommand{\attr}[1]{\mathit{#1}}
\newcommand{\battr}[1]{\mathit{\underline{#1}}}
\newcommand{\rel}[1]{\mathit{#1}}
\newcommand{\ev}[1]{\mathbf{#1}}

\def\punto{\hspace*{\fill}\Box}

\hyphenation{either} \providecommand\AMSLaTeX{AMS\,\LaTeX}
\newcommand\eg{\emph{e.g.}\ }
\newcommand\etc{\emph{etc.}}
\newcommand\bcmdtab{\noindent\bgroup\tabcolsep=0pt%
  \begin{tabular}{@{}p{10pc}@{}p{20pc}@{}}}
\newcommand\ecmdtab{\end{tabular}\egroup}
\newcommand\rch[1]{$\longrightarrow\rlap{$#1$}$\hspace{1em}}
\newcommand\lra{\ensuremath{\quad\longrightarrow\quad}}

  \title[Theory and Practice of Logic Programming]
        {Experimenting with recursive queries in database and logic programming systems}

  \author[G. Terracina, N. Leone, V. Lio, C. Panetta]
         {G. TERRACINA, N. LEONE, V. LIO, C. PANETTA\\
         Dipartimento di Matematica, Universit\`a della Calabria,\\
         Via P. Bucci, 87030, Rende (CS), Italy\\
         \email{\{terracina, leone, lio, panetta\}@mat.unical.it}}

\begin{document}

\label{firstpage}

\maketitle

\begin{abstract}
This paper considers the problem of reasoning on massive amounts of (possibly distributed) data.
Presently, existing proposals show some limitations: {\em (i)} the quantity of data that can be
handled contemporarily is limited, due to the fact that reasoning is generally carried out in
main-memory; {\em (ii)} the interaction with external (and independent) DBMSs is not trivial and,
in several cases, not allowed at all; {\em (iii)} the efficiency of present implementations is
still not sufficient for their utilization in complex reasoning tasks involving massive amounts
of data. This paper provides a contribution in this setting; it presents a new system, called
\dlvdb, which aims to solve these problems. Moreover, the paper reports the results of a thorough
experimental analysis we have carried out for comparing our system with several state-of-the-art
systems (both logic and databases) on some classical deductive problems; the other tested systems
are: LDL++, XSB, Smodels and three top-level commercial DBMSs. \dlvdb significantly outperforms
even the commercial Database Systems on recursive queries.

{\em \noindent To appear in Theory and Practice of Logic Programming (TPLP).}

\end{abstract}

  \begin{keywords}
    Deductive Database Systems, Answer Set Programming / Declarative Logic Programming, Recursive Queries, Benchmarks
  \end{keywords}

\section{Introduction}
\label{sec:Introduction}

The problem of handling massive amounts of data received much attention in the research related
to the development of efficient Database Management Systems (DBMSs).  In this scenario, a
mounting wave of data intensive and knowledge based applications, such as Data Mining, Data
Warehousing and Online Analytical Processing has created a strong demand for more powerful
database languages and systems. An important effort in this direction has been carried out with
the introduction of the latest standard for SQL, namely SQL99 \cite{SQL99} which provides, among
other features, support to object oriented databases and recursive queries.

However, the adoption of SQL99 is still far from being a ``standard''; in fact almost all current
DBMSs do not fully support SQL99 and, in some cases, they adopt proprietary (non standard)
language constructs and functions to implement parts of it. Moreover, the efficiency of current
implementations of SQL99 constructs and their expressiveness are still not sufficient for
performing complex reasoning tasks on huge amounts of data.

The needed expressiveness for reasoning tasks can be provided by logic-based systems. In fact,
declarative logic programming provides a powerful formalism capable of easily modelling and
solving complex problems. The recent development of efficient logic-based systems like \dlv
\cite{leon-etal-2002-dlv}, Smodels \cite{niem-etal-2000a},  XSB \cite{Rao*97}, ASSAT
\cite{lin02a,lin04a}, Cmodels \cite{giu04gs,giu06a}, CLASP \cite{geb07a}, etc., has renewed the
interest in the area of non-monotonic reasoning and declarative logic programming for solving
real world problems in a number of application areas. However, ``data intensive'' problems can
not be handled in a typical logic programming system working in main-memory.

In the past, Deductive Database Systems (DDS) have been proposed to combine the expressive power
of logic-based systems with the efficient data management of DBMSs
\cite{arni-etal-03,Gallaire*84,Ceri*90,GraMin92}; basically, they are an attempt to adapt typical
Datalog systems, which have a ``smalldata'' view of the world, to a ``largedata'' view of the
world via intelligent interactions with some DBMSs. Recently emerging application contexts such
as the ones arising from the natural recursion across nodes in the Internet, or from the success
of intrinsically recursive languages such as XML \cite{Winslett06}, renewed the interest in such
kinds of systems \cite{Abiteboul*05,Loo*05}.

However, the main limitations of currently existing DDSs reside both in the fact that reasoning
is still carried out in main-memory -- this limits the amount of data that can be handled -- and
in the limited interoperability with generic, external, DBMSs they provide. In fact, generally,
the reasoning capabilities of these systems are tailored on a specific (either commercial or
ad-hoc) DBMS.

Summarizing: {\em (i)}  Database systems are nowadays robust and flexible enough to efficiently
handle large amounts of data, possibly distributed; however, their query languages are not
sufficiently expressive to support reasoning tasks on such data. {\em (ii)} Logic-based systems
are endowed with highly expressive languages, allowing them to support complex reasoning tasks,
but they work in main-memory and, hence, can only handle limited amounts of data. {\em (iii)}
Deductive database systems allow to access and manage data stored in DBMSs, however they perform
their computations mainly in main-memory and provide limited interoperability with external (and
possibly distributed) DBMSs.

This work provides a contribution in this setting, bridging the gap between logic-based DDSs and
DBMSs. It presents a new system, named \dlvdb, which is logic-based (like a DDS) but can do all
the work in mass-memory and, in practice, does not have any limitation in the dimension of input
data; moreover, it is capable to exploit optimization techniques both from DBMS (e.g., join
orderings \cite{ullm-etal-00}) and DDS theory (e.g., magic sets \cite{beer-rama-91,Mumick*96}).

\dlvdb allows for two typologies of execution: {\em (i)} direct database execution, which
evaluates logic programs directly on database, with a very limited usage of main-memory but with
some limitations on the expressiveness of the queries, and {\em (ii)} main-memory execution,
which loads input data from different (possibly distributed) databases and executes the logic
program directly in main-memory. In both cases, interoperation with databases is provided by ODBC
connections; these allow handling, in a quite simple way, data residing on various databases over
the network. In order to avoid possible confusion, in the following we use the symbol \dlvdb to
indicate the whole system when the discussion is independent of the execution modality; however,
when it is needed to distinguish between the two execution modalities, we use the symbol \dlvio
to indicate the main-memory execution, whereas the symbol \dlvdb to indicate the direct database
execution.

Summarizing, the overall contributions of this work are the following: {\em (i)}  The development
of a fully fledged system enhancing in different ways the interactions between logic-based
systems and DBMSs.
    {\em (ii)} The development of an efficient, purely database-oriented, evaluation strategy for logic
    programs which minimizes the usage of main-memory and
    maximizes the advantages of optimization techniques implemented
    in existing DBMSs.
    {\em (iii)} The definition of a framework for carrying out an
    experimental comparative analysis of the performance of state-of-the-art systems and \dlvdb.
    {\em (iv)} The execution of a thorough experimentation which shows that \dlvdb beats, often
    with orders of magnitude, Logic-Based Systems (LDL++, XSB, and Smodels\footnote{It is worthwhile noting that,
    since benchmark programs are stratified, they are completely solved by the grounding layer of Smodels (LParse).
    This is the reason why we have not experimented also with ASSAT, Cmodels, and CLASP, as they also use LParse
    for grounding.})
    and even commercial DBMSs
    both for running times and amount of handled data
    on classical deductive problems \cite{ban-ram-88}.

The work is organized as follows. Section \ref{sec:language} presents the reasoning language
supported by the system, whereas Section \ref{sec:Functionalities} describes the functionalities
it provides. In Section \ref{sec:Implementation} the main implementation principles adopted in
the development of \dlvdb are discussed and Section \ref{sec:Architecture} illustrates its
general architecture. Section \ref{sec:experiments} first presents an overview of the
state-of-the-art systems related to \dlvdb, then it describes the experimental analysis we have
carried out to compare \dlvdb with these systems on classical DDS problems. Finally, in Section
\ref{sec:conclusions} we draw our conclusions.

\section{The reasoning language of the system}\label{sec:language}

In this section we briefly describe the syntax and the semantics of the reasoning language
adopted by the \dlvdb system. This is basically Disjunctive Logic Programming (DLP) with
Aggregate functions under the Answer Set Semantics; we refer to this language as \DLPA in the
following. The interested reader can find all details about \DLPA in \cite{fabe-etal-2004-jelia}.

Before starting the presentation, it is worth pointing out that the direct database execution
modality supports only a strict subset of the reasoning language supported by the main-memory
execution. In particular, while \dlvio supports the whole language of \dlv (including
disjunction, unlimited negation, and stratified aggregates), \dlvdb supports or-free programs
with stratified negation and aggregates.

\subsection{Syntax}
We assume that the reader is familiar with standard DLP; we refer to atoms, literals, rules, and
programs of DLP, as {\em standard atoms, standard literals, standard rules}, and {\em standard
programs,} respectively. For further background, see \cite{bara-2002,gelf-lifs-91}.

\vmedio
\paragraph*{Set Terms.}
A (\DLPA) {\em set term} is either a symbolic set or a ground set. A {\em symbolic set} is a pair
$\{\mathit{Vars}\! :\! \mathit{Conj}\}$, where $\mathit{Vars}$ is a list of variables
and $\mathit{Conj}$ is a conjunction of standard atoms.%
\footnote{Intuitively, a symbolic set $\{X \!\! : \!\! a(X,Y), p(Y)\}$ stands for the set of
$X$-values making $a(X,Y), p(Y)$ true, i.e., \mbox{$\{X\! \mid\! \exists Y \mathit{s.t.}\ a(X,Y),
p(Y)\ \mathit{is\ true}\}$}.} A {\em ground set} is a set of pairs of the form
$\tuple{\overline{t}\! :\! \mathit{Conj}}$, where $\overline{t}$ is a list of constants and
$\mathit{Conj}$ is a ground (variable free) conjunction of standard atoms.

\vmedio
\paragraph*{Aggregate Functions.}
An {\em aggregate function} is of the form $f(S)$, where $S$ is a set term, and $f$ is an {\em
aggregate function symbol}. Intuitively, an aggregate function can be thought of as a (possibly
partial) function mapping multisets of constants to a constant.

The aggregate functions which are currently supported are:
\countagg\ (number of terms)%
, \sumagg\ (sum of non-negative rational numbers)%
, \minagg\ (minimum term, undefined for empty set)%
, \maxagg\ (maximum term, undefined for empty set)%
, \avgagg\ (average of non-negative rational numbers)%
\footnote{The first two aggregates correspond, respectively, to the cardinality and weight
constraint literals of Smodels.}.

\vmedio
\paragraph*{Aggregate Literals.}
An {\em aggregate atom} is $f(S) \prec T$, where $f(S)$ is an aggregate function, $\prec \in \{
=,\ <,\ \leq, >, \geq \}$ is a predefined comparison operator, and $T$ is a term (variable or
constant) referred to as guard.

An example of aggregate atom is: $\maxagg\{Z: r(Z), a(Z,V)\} > Y.$

An {\em atom} is either a standard (DLP) atom or an aggregate atom. A {\em literal} $L$ is an
atom $A$ or an atom $A$ preceded by the default negation symbol \nafsymbol; if $A$ is an
aggregate atom, $L$ is an {\em aggregate literal}.

\vmedio
\paragraph*{\DLPA Programs.}
A {\em \DLPA rule} \R{} is a construct
\vpiccolo
\begin{dlvcode}
a_1\Or\cdots\Or a_n \derives\
        b_1,\cdots, b_k,\
        \naf\ b_{k+1},\cdots,\ \naf\ b_m.
\end{dlvcode}
\vgrande

\noindent where $a_1,\ldots,a_n$ are standard atoms, $b_1,\cdots ,b_m$ are atoms, $n\geq 0$, and
$m\geq k\geq 0$. The disjunction $a_1\Or\cdots\Or a_n$ is referred to as the {\em head} of \R{}
whereas the conjunction $b_1,...,b_k,\ \naf\ b_{k+1},$ $..., \naf\ b_m$ is the {\em body} of
\R{}. We denote the set $\{a_1,\ldots,a_n\}$ of the head atoms by $H(\R{})$, and the set
$\{b_1,...,b_k,\ \naf\ b_{k+1},..., \naf\ b_m\}$ of the body literals by $B(\R{})$. $B^{+}(\R{})$
and $B^{-}(\R{})$ denote, respectively, the set of positive literals and the set of negative
literals occurring in $B(\R{})$.

A {\em \DLPA program} \aprog\ is a set of \DLPA rules.

Note that \DLPA allows also for built-in predicates \cite{dlv-web} in its rules, such as the
comparative predicates equality, less-than, and greater-than (\code{=}, $<$, $>$) and arithmetic
predicates like addition or multiplication (+, *).

\vpiccolo
\paragraph*{Safety.} A {\em global} variable of a rule $r$ is a
variable appearing in a standard atom of $r$; all other variables are {\em local} variables. A
rule $r$ is {\em safe} if the following conditions hold: {\em (i)} each global variable of $r$
appears in a positive standard literal in the body of $r$; {\em (ii)} each local variable of $r$
appearing in a symbolic set $\{\mathit{Vars}:\mathit{Conj}\}$ appears in an atom of
$\mathit{Conj}$; {\em (iii)} each guard of an aggregate atom of $r$ is a constant or a global
variable. A program $\p$ is safe if all $\R \in \p$ are safe. In the following we assume that
\DLPA programs are safe.

$\ $ \\
Let the {\em level mapping} of a program $\aprog{}$ be a function $||\ ||$ from the predicates in
$\aprog{}$ to finite ordinals; moreover, given an atom $A=p(t_1,\ldots ,t_n)$, we denote by
$\mathit{Pred}(A)$ its predicate $p$.

\vmedio
\paragraph*{Stratified$^{not}$ programs.}
A {\em \DLPA program}  $\aprog{}$ is called {\em stratified$^{not}$} \cite{apt-etal-88,przy-88},
if there is a level mapping $||\ ||_s$ of $\aprog{}$ such that, for every rule $r$: (1) for any
$l \in \BpR$, and for any $l' \in H(r)$, $||Pred(l)||_s \leq ||Pred(l')||_s$; (2) for any $l \in
\BnR$, and for any $l' \in H(r)$, $||Pred(l)||_s < ||Pred(l')||_s$; (3) for any $l, l' \in H(r)$,
$||Pred(l)||_s = ||Pred(l')||_s$.

\vpiccolo
\paragraph*{Stratified$^{aggr}$ programs.}
A {\em \DLPA program}  $\aprog{}$ is called {\em stratified$^{aggr}$} \cite{dell-etal-2003b}, if
there is a level mapping $||\ ||_a$ of $\aprog{}$ such that, for every rule $r$: (1)  if $l$
appears in the head of $r$, and $l'$ appears in an aggregate atom in the body of $r$, then
$||Pred(l')||_a < ||Pred(l)||_a$; and (2) if $l$ appears in the head of $r$, and $l'$ occurs in a
standard atom in the body of $r$, then $||Pred(l')||_a\leq ||Pred(l)||_a$. (3) If both $l$ and
$l'$ appear in the head of $r$, then $||Pred(l')||_a = ||Pred(l)||_a$.

\begin{example}\label{ex:stratification}
Consider the program consisting of a set of facts for predicates $a$ and $b$,
plus the following two rules:\\
\vmedio
\begin{dlvcode}
q(X) \derives p(X), \countagg\{Y : a(Y,X),b(X)\} \leq 2. \quad \quad
p(X) \derives q(X), b(X).\\
\end{dlvcode}
The program is stratified$^{aggr}$, as the level mapping \ \ $||a||=||b||=1$, $||p||=||q||=2$ \ \
satisfies the required conditions. If we add the rule $b(X) \derives p(X)$, then no level-mapping
exists and the program becomes not stratified$^{aggr}$. $\punto$ \end{example}

\noindent Intuitively, the property stratified$^{aggr}$ forbids recursion through aggregates.

\vmedio
\paragraph*{Supported languages.}
The direct database execution modality (\dlvdb) currently supports only {\em \DLPA} programs
which are disjunction free, stratified$^{not}$, and strat-ified$^{aggr}$. Note that both built-in
predicates and aggregates are supported.

Conversely, the main-memory execution modality (\dlvio) supports all {\em \DLPA} programs that
are stratified$^{aggr}$. As a consequence, unrestricted negation, disjunction and non recursive
aggregates are supported.

\subsection{Answer Set Semantics}
\label{sub:aspsemantics}

\paragraph*{Universe and Base.}
Given a \DLPA program \p, let \UP\ denote the set of constants appearing in \p, and \BP\ be the
set of standard atoms constructible from the (standard) predicates of \p\ with constants in \UP.
Given a set $X$, let $\multi{X}$ denote the set of all multisets over elements from $X$. Without
loss of generality, we assume that aggregate functions  map to $\rational$ (the set of rational
numbers).

\vpiccolo
\paragraph*{Instantiation.}
A {\em substitution} is a mapping from a set of variables to $\UP$. A substitution from the set
of global variables of a rule $r$ (to $\UP$) is a {\em global substitution for r}; a substitution
from the set of local variables of a symbolic set $S$ (to $\UP$) is a {\em local substitution for
$S$}. Given a symbolic set without global variables $S = \{\!\mathit{Vars}:\mathit{Conj}\}$, the
\emph{instantiation of $S$} is the following ground set of pairs $inst(S)$: $\{
\tuple{\gamma(\mathit{Vars}): \gamma(\mathit{Conj})} \mid$
$\gamma$ {\em is a local substitution for }$S\}$.%
\footnote{Given a substitution $\sigma$ and a \DLPA object $Obj$ (rule, set, etc.), we denote by
$\sigma(Obj)$
the object obtained by replacing each variable $X$ in $Obj$ by $\sigma(X)$.}\\
A {\em ground instance} of a rule $r$ is obtained in two steps: (1) a global substitution
$\sigma$ for $r$ is first applied over $r$; (2) every symbolic set $S$ in $\sigma(r)$ is replaced
by its instantiation $inst(S)$. The instantiation \GP\ of a program \p\ is the set of all
possible instances of the rules of \p.
\begin{example} \em
{\em \label{ex:instantiation} Consider the following program $\p_1$:
\vpiccolo
\begin{dlvcode}
q(1) \Or p(2,2).  \quad \quad q(2) \Or p(2,1).  \quad \quad
t(X) \derives q(X),
\sumagg\{Y:p(X,Y)\}>1.
\end{dlvcode}
The instantiation $\ground{\p_1}$ is the following:
\vpiccolo
\begin{dlvcode}
q(1) \Or p(2,2). \quad \quad
t(1) \derives q(1), \sumagg\{\tuple{1\!:\!p(1,1)},\tuple{2\!:\!p(1,2)}\}\!>\!1. \\
q(2) \Or p(2,1).\quad \quad t(2) \derives q(2),
\sumagg\{\tuple{1\!:\!p(2,1)},\tuple{2\!:\!p(2,2)}\}\!>\!1.
\vspace*{-0.7cm} \end{dlvcode}
$\punto$
} 
\vmedio

\end{example}

\paragraph*{Interpretations.}
 An {\em interpretation} for a \DLPA program \p\ is a  set
of standard ground atoms, that is $I\subseteq \BP$. A positive literal $A$ is true w.r.t. $I$ if
$A \in I$, is false otherwise. A negative literal $\naf\ A$ is true w.r.t. $I$, if $A \not\in I$,
it is false otherwise.

An interpretation also provides a meaning for aggregate literals.

Let $I$ be an interpretation. A standard ground conjunction is true (resp. false) w.r.t. $I$ if
all its literals are true. The meaning of a set, an aggregate function, and an aggregate atom
under an interpretation, is a multiset, a value, and a truth-value, respectively. Let $f(S)$ be a
an aggregate function. The valuation $I(S)$ of $S$ w.r.t.\ $I$ is the multiset of the first
constant of the elements in $S$ whose conjunction is true w.r.t.\ $I$. More precisely, let $I(S)$
denote the multiset $[ t_1 \mid \tuple{t_1,...,t_n\!:\!\mathit{Conj}}\!\in\! S \wedge$ {\em
$\mathit{Conj}$ is true w.r.t.\ I}$\ ]$. The valuation $I(f(S))$ of an aggregate function $f(S)$
w.r.t.\ $I$ is the result of the application of $f$ on $I(S)$. If the multiset $I(S)$ is not in
the domain of $f$, $I(f(S))= \bot$ (where $\bot$ is a fixed symbol not
occurring in \p).%

An instantiated aggregate atom $A = f(S) \prec k$ is {\em true w.r.t.\ $I$} if: {\em (i)}
$I(f(S))\neq \bot$, and, {\em (ii)}  $I(f(S)) \prec k$ holds; otherwise, $A$ is false. An
instantiated aggregate literal $\naf\ A = \naf (f(S) \prec k)$ is {\em true w.r.t.\ $I$} if: {\em
(i)} $I(f(S))\neq \bot$, and, {\em (ii)}  $I(f(S)) \prec k$ does not hold; otherwise, $A$ is
false.

\vpiccolo
\paragraph*{Minimal Models.}
Given an interpretation $I$, a rule $r$ is {\em satisfied w.r.t.\ $I$} if some head atom is true
w.r.t.\ $I$ whenever all body literals are true w.r.t.\ $I$. An interpretation $M$ is a {\em
model} of a \DLPA program $\p$ if all $\R \in \GP$ are satisfied w.r.t.\ $M$. A model $M$ for
$\p$ is (subset) minimal if no model $N$ for $\p$ exists such that $N \subset M$.

\vpiccolo
\paragraph*{Answer Sets.}
We now recall the generalization of the Gelfond-Lifschitz transformation to programs with
aggregates from \cite{fabe-etal-2004-jelia}.

\begin{definition}[\cite{fabe-etal-2004-jelia}]\label{def:reduct}
{\em Given a ground \DLPA program $\p$ and a total interpretation $I$, let $\p^I$ denote the
transformed program obtained from $\p$ by deleting all rules in which a body literal is false
w.r.t.\ $I$. $I$ is an answer set of a program $\p$ if it is a minimal model of $\GP^I$. }
\end{definition}

\begin{example} \em \label{ex:simple-trans}
{\em Consider the following two programs:
\vpiccolo
\begin{dlvcode}
P_1:  \{p(a) \derives \countagg\{X: p(X)\}>0.\} \quad \quad
P_2:  \{p(a) \derives \countagg\{X: p(X)\}<1.\}
\end{dlvcode}
\vmedio

\noindent $\ground{P_1} =$ $\{p(a) \derives \countagg\{\tuple{a: p(a)}\}>0.\}$ \ and
$\ground{P_2} =\{p(a) $\  $\derives \countagg\{\tuple{a:%
p(a)}\}<1.\}$; consider also interpretations $I_1 = \{p(a)\}$ and $I_2 = \emptyset$. Then,
$\ground{P_1}^{I_1} = \ground{P_1}$, $\ground{P_1}^{I_2} = \emptyset$, and $\ground{P_2}^{I_1} =
\emptyset$, $\ground{P_2}^{I_2} = \ground{P_2}$ hold. $I_2$ is the only answer set of $P_1$
(because $I_1$ is not a minimal model of $\ground{P_1}^{I_1}$), whereas $P_2$ admits no answer
set ($I_1$ is not a minimal model of $Ground(P_2)^{I_1}$, and $I_2$ is not a model of
$Ground(P_2) = Ground(P_2)^{I_2}$).
} 
$\punto$
\end{example}

Note that any answer set $A$ of $\p$ is also a model of $\p$ because $\GP^A \subseteq \GP$, and
rules in $\GP - \GP^A$ are satisfied w.r.t.\ $A$.

\section{System Functionalities}
\label{sec:Functionalities}

As pointed out in the Introduction, the presented system allows for two typologies of execution:
{\em (i)} direct database execution (\dlvdb), which is capable of handling massive amounts of
data but with some limitations on the expressiveness of the query program (see Section
\ref{sec:language}), and {\em (ii)} main-memory execution (\dlvio) which allows the user to take
full advantage of the expressiveness of \DLPA and to import data residing on DBMSs, but with some
limitations on the quantity of data to reason about, given by the amount of available
main-memory.

The system, along with a manual and some examples, is available for download at the address
{\small {\tt http://www.mat.unical.it/terracina/dlvdb}}. In the following we provide a general
description of the main functionalities provided by \dlvdb and \dlvio. The interested reader can
find all details on the system's web site.

\subsection{Direct Database Execution}

Three main peculiarities characterize the \dlvdb system in this execution modality: {\em (i)} its
ability to evaluate logic programs directly and completely on databases with a very limited usage
of main-memory resources, {\em (ii)} its capability to map program predicates to (possibly
complex and distributed) database views, and {\em (iii)} the possibility to easily specify which
data is to be considered as input or as output for the program. This is the main contribution of
our work.

Roughly speaking, in this execution modality the user has his data stored in (possibly
distributed) database tables and wants to carry out some reasoning on them; however the amount of
such data, or the number of facts that are generated during the reasoning, is such that the
evaluation can not be carried out in main-memory. Then, the program must be evaluated directly in
mass-memory.

In order to properly carry out the evaluation, it is necessary to specify the mappings between
input and output data and program predicates, as well as to specify wether the temporary
relations possibly needed for the mass-memory evaluation should be maintained or deleted at the
end of the execution. These can be specified by some auxiliary directives. The grammar in which
these directives must be expressed is shown in Figure \ref{grammar}.

\begin{figure*}[t]
{\scriptsize \begin{verbatim}
 Auxiliary-Directives ::= Init-section [Table-definition]+ [Query-Section]? [Final-section]*
 Init-Section ::=USEDB DatabaseName:UserName:Password [System-Like]?.
 Table-definition ::=
    [USE TableName [( AttrName [, AttrName]* )]? [AS ( SQL-Statement )]?
    [FROM DatabaseName:UserName:Password]?
    [MAPTO PredName [( SqlType [, SqlType]* )]? ]?.
    |
    CREATE TableName [( AttrName [, AttrName]* )]?
    [MAPTO PredName [( SqlType [, SqlType]* )]? ]?
    [KEEP_AFTER_EXECUTION]?.]
 Query-Section ::= QUERY TableName.
 Final-section ::=
    [DBOUTPUT DatabaseName:UserName:Password.
    |
    OUTPUT [APPEND | OVERWRITE]? PredName [AS AliasName]?
    [IN DatabaseName:UserName:Password.]
 System-Like ::= LIKE [POSTGRES | ORACLE | DB2 | SQLSERVER | MYSQL]
\end{verbatim}}

 \caption{Grammar of the auxiliary directives.} \label{grammar}
\end{figure*}

Intuitively, the user must specify the working database in which the system has to perform the
evaluation. Moreover, he can specify a set of table definitions; note that each specified table
is mapped into one of the program predicates. Facts can reside on separate databases or they can
be obtained as views on different tables. Attribute type declaration is needed only for a correct
management of built-in predicates. The {\tt USE} and {\tt CREATE} options can be exploited to
specify input and output data as well as temporary relations needed for the mass memory
instantiation. Finally, the user can choose to copy the entire output of the evaluation or parts
thereof in different databases.

\begin{example} \label{example}
Assume that a travel agency asks to derive all the destinations reachable by an airline company
either by using its aircrafts or by exploiting code-share agreements. Suppose that the direct
flights of each company are stored in a relation {\tt flight\_rel(Id, FromX, ToY, Company)} of
the database {\tt dbAirports}, whereas the code-share agreements between companies are stored in
a relation {\tt codeshare\_rel} {\tt(Company1, Company2, FlightId)} of an external database {\tt
dbCommercial}; if a code-share agreement holds between the company $c1$ and the company $c2$ for
$flightId$, it means that the flight $flightId$ is actually provided by an aircraft of $c1$ but
can be considered also carried out by $c2$. Finally, assume that, for security reasons, travel
agencies are not allowed to directly access the databases {\tt dbAirports} and {\tt
dbCommercial}, and, consequently, it is necessary to store the output result in a relation {\tt
composedCompanyRoutes} of a separate database {\tt dbTravelAgency} supposed to support travel
agencies. The \DLPA program that can derive all the connections is:

\[{\small
\begin{array}{lll}
(1) & destinations(FromX, ToY, Comp)\ \derives\ & flight(Id, FromX, ToY, Comp).\\
(2) & destinations(FromX, ToY, Comp)\ \derives\ & flight(Id, FromX, ToY, C2), \\
    &                                         & codeshare(C2,Comp, Id).\\
(3) & destinations(FromX, ToY, Comp)\ \derives\ & destinations(FromX, T2, Comp), \\
&& destinations(T2, ToY, Comp).
\end{array}}
\]

\noindent In order to exploit data residing in the above mentioned databases, we should map the
predicate $flight$ to the relation {\tt flight\_rel} of {\tt dbAirports} and the predicate
$codeshare$ to the relation {\tt codeshare\_rel} of {\tt dbCommercial}. Finally, we have to map
the predicate $destinations$ to the relation {\tt composedCompanyRoutes} of {\tt dbTravelAgency}.

Now suppose that, due to a huge size of input data, we need to evaluate the program in
mass-memory (on a DBMS). In order to carry out this task, the auxiliary directives shown in
Figure \ref{fig:mappings} should be used. They allow to specify the mappings between the program
predicates and the database relations introduced previously. $\punto$

\begin{figure}
\vspace*{-0.0cm} {\scriptsize\begin{verbatim}
 USEDB dlvdb:myname:mypasswd.
 USE flight_rel (Id, FromX, ToY, Company) FROM dbAirports:airportUser:airportPasswd
 MAPTO flight (integer, varchar(255), varchar(255), varchar(255)).
 USE codeshare_rel (Company1, Company2, FlightId) FROM dbCommercial:commUser:commPasswd
 MAPTO codeshare (varchar(255), varchar(255), integer).
 CREATE destinations_rel (FromX, ToY, Company)
 MAPTO destinations (varchar(255), varchar(255), varchar(255)) KEEP_AFTER_EXECUTION.
 OUTPUT  destinations AS composedCompanyRoutes IN dbTravelAgency:agencyName:agencyPasswd.
\end{verbatim}}
 \caption{Auxiliary directives for Example \ref{example2}.} \label{fig:mappings}
\end{figure}

\end{example}

It is worth pointing out that if a predicate is not explicitly mapped into a table, but a
relation with the same name and compatible attributes is present in the working database, the
system automatically hypothesize a {\tt USE} mapping for them. Analogously, if a predicate is not
explicitly mapped and no corresponding table exists in the working database, a {\tt CREATE}
mapping is automatically hypothesized for it. This significantly simplifies the specification of
the auxiliary directives; in fact, in the ideal case -- when everything is in the working
database and each input predicate has the corresponding input table with the same name -- only
the {\tt Init-Section} and one of {\tt CREATE} or {\tt OUTPUT} options are actually needed to run
a program and check its output.

\subsection{Main-Memory Execution}

The main-memory execution modality of the system  allows input facts to be (possibly complex)
views on database tables and allows exporting (parts of) predicates to database relations.
However, the program evaluation is carried out completely in main-memory; this allows the system
to evaluate more complex logic programs (see Section \ref{sec:language}) but at the price of a
lower amount of data the system can handle, due to the limited amount of main-memory.

The concept of importing and exporting data from external data sources into logic-based systems
is not new (see, for example \cite{arni-etal-03,Lu*96,Rao*97}); the contribution of this
execution modality is mainly of technological relevance and has the merit of providing Answer Set
Programming with an easy way to access distributed data spread over the network. Another
advancement w.r.t. existing proposals is its flexibility in the types of external source that can
be accessed; in fact, most of the existing systems are tailored on custom DBMSs, whereas our
system can be interfaced with any external source which provides an ODBC connection.

Intuitively, \dlvio can be exploited when the user has to perform very complex reasoning tasks
(in the \NP class or higher) but the data is available in database relations, or the output must
be permanently stored in a database for further elaborations.

In order to perform these tasks, two built-in commands are added in \dlvio to the standard \DLPA
syntax, namely the \#import and the \#export commands:

\begin{quote}
{\small \#import(databasename,``username'',``password'',``query'',predname, typeConv).

\#export(databasename,``username'',``password'',predname,tablename).}
\end{quote}

An \#import command retrieves data from a table ``row by row'' through the \emph{query} specified
by the user in SQL and creates one atom for each selected tuple. The name of each imported atom
is set to {\em predname}, and is considered as a fact of the program.

The \#export command generates a new tuple into {\em tablename} for each new truth value derived
for {\em predname} by the program evaluation.

An alternative form of the \#export command is the following:

\begin{quote}
\#export(databasename, ``username'', ``password'', predname, tablename,

\hfill{``REPLACE where $<$condition$>$'' )}
\end{quote}

\noindent which can be used to remove from {\em tablename} the tuples of {\em predname} for which
the ``REPLACE where'' condition holds; it can be useful for deleting tuples corresponding to
violated integrity constraints.

It is worth pointing out that if a \DLPA program contains at least one \#export command, the
system can compute only the first valid answer set; this limitation has been introduced mainly to
avoid an exponential space complexity of the system. In fact, the number of answer sets can be
exponential in the input.

\begin{example}
\label{example2}

Consider again the scenario introduced in Example \ref{example}, and assume that the amount of
input data allows the evaluation to be carried out in main-memory. The built-in commands that
must be added to the \DLPA program of Example \ref{example} to implement the necessary mappings
are:

{\small \noindent \#import(dbAirports, ``airportUser'', ``airportPasswd'' , ``SELECT * FROM
flight\_rel'',

\hfill{flight, type : U\_INT, Q\_CONST, Q\_CONST, Q\_CONST).}

\noindent \#import(dbCommercial, ``commUser'', ``commPasswd'', ``SELECT * FROM codeshare\_rel'',

\hfill{codeshare, type : Q\_CONST, Q\_CONST, U\_INT).}

\noindent \#export(dbTravelAgency, ``agencyName'', ``agencyPasswd'', destinations,

\hfill{composedCompanyRoutes). }}

$\punto$
\end{example}

Note that the syntax of \dlvio directives is simpler than that of \dlvdb auxiliary directives.
This is because \dlvio is intended to provide an easy mechanism to load data into the logic
program and then store its results back to mass-memory, whereas \dlvdb is oriented to more
sophisticated applications handling distributed data and mass-memory-based reasoning and,
consequently, it must provide a richer set of options in defining the mappings.

\section{Implementation principles}
\label{sec:Implementation}

The main innovation of our system resides in the evaluation of \DLPA programs directly on a
database. The evaluation process basically consists of two steps: {\em (i)} the translation of
\DLPA rules in SQL statements, {\em (ii)} the definition of an efficient SQL query plan such that
the computed answers are the same as the ones of the main-memory execution, but where the
evaluation process is completely carried out in mass-memory. In the following, we first describe
the general philosophy of our mass-memory evaluation strategy, then we present the algorithms
used to obtain SQL statements from \DLPA rules.

\subsection{General characteristics of the evaluation strategy}
\label{sub:evaluation}

The evaluation of a program $\cal P$ starts from the analysis of its intensional component. In
particular, $\cal P$ is first transformed into an equivalent program $\cal P'$ which can be
evaluated more efficiently by the subsequent steps. Transformations carried out in this phase
take into account various aspects of the input program; as an example, they aim to {\em (i)}
reduce the arity of intermediate relations whenever possible, {\em (ii)} reduce the size of
intermediate relations \cite{fabe-etal-99c}, {\em (iii)} push down constants in the queries by
magic-sets rewritings \cite{Bancilhon*86,beer-rama-91,Mumick*96,Ross90}, etc. All these
optimizations do not take into account the extensional component (the facts) of $\cal P$; some
other optimizations are described in \cite{fabe-etal-99c}.

After this, the connected components and their topological order (i.e., the Dependency Graph) of
the resulting program are computed. Then, it is evaluated one component at a time, starting from
the lowest ones in the topological order.

The evaluation of each component follows the Semi-Naive method \cite{ullm-89} with the
enhancements showed in \cite{bal-ram-87,zani-etal-97} to optimize the evaluation of rules with
non-linear recursion.

In particular, the Semi-Naive algorithm applied to a component ${\cal P}_{C}$ can be viewed as a
two-phase algorithm: the first one deals with non-recursive rules, which can be completely
evaluated in one single step; the second one deals with recursive rules which need an iterative
fixpoint computation for their complete evaluation. At each iteration there are a number of
predicates whose extensions have been already fully determined (predicates not belonging to
${\cal P}_{C}$ which have been therefore previously evaluated), and a number of recursive
predicates (i.e., belonging to ${\cal P}_{C}$) for which a new set of truth values can be
determined from the available ones. Then, in order to evaluate, e.g., the rule:

\begin{quote}
$ (r_1): \quad   p_{0}(X, Y) \derives p_{1}(X, Y), p_{2}(Y, Z), q(X, Z).$
\end{quote}

\noindent where $p_{1}$ and $p_{2}$ are mutually recursive with $p_{0}$ and $q$ is not recursive,
the standard Semi-Naive method evaluates the following formula (expressed in relational algebra)
at each iteration:

\begin{quote} $
 \begin{array}{lccccccc}

 \Delta P_{0}^{k} = & \Delta P_{1}^{k-1} & \bowtie & P_{2}^{k-1} & \bowtie & Q & \cup & (a)\\
 & P_{1}^{k-1} & \bowtie & \Delta P_{2}^{k-1} & \bowtie & Q &   & (b)

 \end{array} $
 \end{quote}

\noindent Here, a capital letter is used to indicate the database relation corresponding to the
(lower case) predicate; $P_{j}^{k}$ indicates the values stored in relation $P_{i}$ up to step
$k$ and $\Delta P_{j}^{k}$ is the set of new values determined for $P_{j}$ at step $k$ (in the
following, we call $\Delta P_{j}^{k}$ the \emph{differential} of $P_{j}$).

However, the standard Semi-Naive approach is characterized by inefficiencies in evaluating
non-linear recursive rules. In fact, if each $P_{j}^{k-1}$ is expanded in its (disjoint)
components $P_{j}^{k-2}$ and $\Delta P_{j}^{k-1}$ the formula $(a) \cup (b)$ above becomes:

\vspace{0.2cm}
\begin{quote} $
\begin{array}{cc}
\begin{array}{l}
\Delta P_{0}^{k} =\\ \\
\end{array}
& \emph{(a)} \left [
\begin{array}{cccccccc}
& \Delta P_{1}^{k-1} & \bowtie & P_{2}^{k-2} & \bowtie & Q & \cup & \emph{ (1)}\\
& \Delta P_{1}^{k-1} & \bowtie & \Delta P_{2}^{k-1} & \bowtie & Q & \cup & \emph{ (2)}
\end{array}
\right . \\
& \emph{(b)} \left [
\begin{array}{cccccccc}
& P_{1}^{k-2} & \bowtie & \Delta P_{2}^{k-1} & \bowtie & Q & \cup & \emph{ (3)} \\
& \Delta P_{1}^{k-1} & \bowtie & \Delta P_{2}^{k-1} & \bowtie & Q &   & \emph{ (4)}
\end{array}
\right . \\
\end{array}
$
\end{quote}

\noindent where {\em (a)} expands $P_{2}^{k-1}$ and {\em (b)} expands $P_{1}^{k-1}$; note that
line {\em (2)} and {\em (4)} are identical. The enhancement described in
\cite{bal-ram-87,zani-etal-97} provides a solution to this problem rewriting the original rule
in:

\vspace{0.2cm}
\begin{quote} $
\begin{array}{cc}
\begin{array}{l}
\Delta P_{0}^{k} =\\ \\
\end{array}
&
\begin{array}{ccccccc}
& \Delta P_{1}^{k-1} & \bowtie & P_{2}^{k-1} & \bowtie & Q & \cup \\
& P_{1}^{k-2} & \bowtie & \Delta P_{2}^{k-1} & \bowtie & Q &
\end{array}
\end{array}
$
\end{quote}

\noindent which, indeed, avoids to re-compute joins in {\em (2)}, {\em (4)} more times.

Generalizing the solution to a rule having $r$ predicates mutually recursive with its head, the
differentiation is obtained by subdividing the original rule in $r$ sub-rules such that the
$i$-th sub-rule has the form $\Delta p_{0}^{k} :- p_{1}^{k-2}, \ldots, p_{i-1}^{k-2}, \Delta
p_{i}^{k-1}, p_{i+1}^{k-1}, \ldots,$ $p_{r}^{k-1}, q$. Note that both relations $P_{j}^{k-1}$,
$P_{j}^{k-2}$ and $\Delta P_{j}^{k-1}$ are considered.

In our approach, we follow a slightly different strategy, which both unfolds {\em each} relation
$P_{j}^{k-1}$ in $P_{j}^{k-2}$ and $\Delta P_{j}^{k-1}$ and avoids to produce the redundant
sub-rules of the standard Semi-Naive method. This is carried out as follows. Let us tag the
differential relations ($\Delta P_{j}^{k-1}$) with the symbol 1 and the standard ones
($P_{j}^{k-2}$) with the symbol 0. Given a generic rule with $r$ predicates $p_1 ,\ldots, p_r$ in
its body mutually recursive with the head, our approach follows the binary enumeration between 1
and $2^{r}-1$ and, for each of these binary numbers, it generates a differential rule; in
particular, if position $j$ on the binary number contains a $0$, then $P_{j}^{k-2}$ is put in the
corresponding rule, otherwise $\Delta P_{j}^{k-1}$ is used. As for the previous example, rule
$(r_1)$ is evaluated, in our approach, with joins {\em (1)}, {\em (2)} and {\em (3)} shown above.

Note that this approach generates a higher number of auxiliary rules w.r.t.
\cite{bal-ram-87,zani-etal-97} but, while avoiding to execute the same set of redundant joins, it
allows handling smaller relations. This could constitute a good advantage when handling massive
amounts of data, because managing several small joins can be less resource demanding than
executing few big ones.

The algorithm implemented in our system for the differential semi-naive evaluation strategy
described above is shown in Figure \ref{alg:semi-naive-mod}. It is executed for each component
$\P_{C}$ of the input program $\P$ and assumes that input \DLPA rules have been already
translated to SQL statements. Here, the component $\P_{C}$ depends on predicates
     $r_{1},\ldots,r_{n}$ solved in previous components and has $q_{1},\ldots,q_{m}$ as non
     recursive predicates or facts and $p_{1},\ldots,p_{n}$ as recursive predicates.

\begin{figure}
%
{\tiny
\begin{tabular}{l}
    \hline
    \hspace*{0cm}{\bf{Differential Semi-Naive}(Input: $R_{1},\ldots,R_{l}$. Output: $Q_{1},\ldots,Q_{m}$, $P_{1},\ldots,P_{n}$)} \\
    \hspace*{0cm}{\kbegin}  \\
    \hspace*{0.5cm}{\kfor\ i:=1 \kto\ m \kdo \ // Evaluate non recursive predicates} \\
    (1) \hspace*{1cm}{$Q_{i} =\mathit{EVAL}(q_i,\ R_{1},\ldots,\ R_{l}, \ Q_{1},\ldots,\ Q_{m})$;} \\
    \hspace*{0.5cm}{\kfor\ i:=1 \kto\ n \kdo \ \kbegin \ // Initialize recursive predicates}  \\
    (2)\hspace*{1cm}{$P_{i}^{k-2} = \mathit{EVAL}(p_i,\ R_{1},\ldots,\ R_{l},\ Q_{1},\ldots,\ Q_{m}) $;} \\
    (3)\hspace*{1cm}{$\Delta P_{i}^{k-1} = P_{i}^{k-2}$;} \\
    \hspace*{0.5cm}{\kend;} \\
    \hspace*{0.5cm}{\krepeat} \\
    \hspace*{1cm}{\kfor\ i:=1 \kto\ n \kdo \ \kbegin} \\
    (4)\hspace*{1.5cm}{$\Delta P_{i}^{k} = \mathit{EVAL\_DIFF}(p_i, P_{1}^{k-2}, \ldots,P_{n}^{k-2}, \Delta
    P_{1}^{k-1},\ldots,\Delta P_{n}^{k-1},R_{1},\ldots,\ R_{l},$$Q_{1},\ldots,\ Q_{m})$;}\\
    (5)\hspace*{1.5cm}{$\Delta P_{i}^{k} = \Delta P_{i}^{k} - P_{i}^{k-2} - \Delta P_{i}^{k-1}$;} \\
    \hspace*{1.0cm}{\kend;} \\
    \hspace*{1cm}{\kfor\ i:=1 \kto\ n \kdo \ \kbegin} \\
    (6)\hspace*{1.5cm}{$P_{i}^{k-2} = P_{i}^{k-2} \cup \Delta P_{i}^{k-1}$;} \\
    (7)\hspace*{1.5cm}{$\Delta P_{i}^{k-1} = \Delta P_{i}^{k}$;} \\
    \hspace*{1.0cm}{\kend;} \\

    \hspace*{0.5cm}{\kuntil\ $\Delta P_{i}^{k} = \emptyset, \forall i\ 1\leq i \leq n$;} \\

    \hspace*{0.5cm}{\kfor\ i:=1 \kto\ n \kdo} \\
    (8)\hspace*{1cm}{$P_{i}= P_{i}^{k-2} $;} \\

    \hspace*{0cm}{\kend .} \\
    \hline
\end{tabular} }

\caption{Algorithm Differential Semi-Naive}\label{alg:semi-naive-mod}
\end{figure}

Function $\mathit{EVAL}(q_i,\ R_{1},\ldots,\ R_{l}, \ Q_{1},\ldots,\ Q_{m})$ performs the
evaluation of the non recursive rules having $q_i$ as head as follows: it first runs each SQL
query corresponding to a rule having $q_i$ as head; then, the corresponding results are added to
the relation $Q_i$.

Function $\mathit{EVAL\_DIFF}(p_i, P_{1}^{k-2}, \ldots,P_{n}^{k-2},$ $\Delta
P_{1}^{k-1},\ldots,\Delta P_{n}^{k-1},R_{1},\ldots, R_{l},Q_{1},\ldots,$ $\ Q_{m})$ implements
the optimization to the Semi-Naive method; it computes the new values for the predicate $p_i$ at
the current iteration $k$ starting from the values computed until iteration $k-2$ and the new
values obtained at the previous iteration $k-1$. In more detail, the SQL statements corresponding
to each recursive rule having $p_i$ as head are considered. The final result of
$\mathit{EVAL\_DIFF}$ is stored in table $\Delta P_{i}^{k}$. Clearly, it cannot be proved that
$\mathit{EVAL\_DIFF}$ does not recompute some truth values already obtained in previous
iterations. As a consequence, $\Delta P_{i}^{k}$ must be cleaned up from these values after the
computation of $\mathit{EVAL\_DIFF}$; this is exactly what is done by instruction (5) of the
algorithm. Instruction (6) and (7) are needed to reuse the same relations ($\Delta P_{i}^{k}$,
$\Delta P_{i}^{k-1}$, $P_{i}^{k-2}$) at each iteration.

Finally, it is worth pointing out that the last \kfor\ of the algorithm (instruction (8)) is
shown just for clarity of exposition; in fact, in the actual implementation, what we indicated as
$P_{i}^{k-2}$ is exactly table $P_{i}$.

It is worth pointing out that the basic step of the evaluation is the execution of standard SQL
queries over the underlying data. In fact, one of the main objectives in the implementation of
\dlvdb has been that of associating one single (non recursive) SQL statement with each rule of
the program (either recursive or not), without the support of main-memory data structures for the
evaluation. This allows \dlvdb to minimize the ``out of memory'' problems caused by limited
main-memory dimensions. Moreover, the overall organization of the evaluation strategy allows
benefiting from both the optimizations on the intensional component of the program (the program
rewriting techniques outlined at the beginning of this section) and the optimizations on the
extensional component (the data) already implemented in the DBMS configured as the working
database.

The combination of such optimizations, along with a wise translation of datalog rules in
efficient SQL queries allow \dlvdb to boost the evaluation process even w.r.t. main-memory
evaluation strategies (see Section \ref{sec:experiments}).

\subsection{From \DLPA to SQL}
\label{sub:DLPA2SQL}

In this section we describe the general functions exploited to translate \DLPA rules in SQL
statements. Functions are presented in pseudocode and, for the sake of presentation clarity, they
omit some details; moreover, since there is a one-to-one correspondence between the predicates in
the logic program and the relations in the database, in the following, when this is not
confusing, we use the terms predicate and relation interchangeably. It is worth recalling that
these one-to-one correspondences are determined both from the user specifications in the
auxiliary directives and from the mappings automatically derived by the system.

In order to provide examples for the presented functions, we exploit the following reference
schema:
\[
\begin{array}{ll}
\rel{employee}(\attr{Ename},\attr{Salary},\attr{Dep},\attr{Boss}) &
\rel{department}(\attr{Code},\attr{Director})
\end{array}
\]
storing information about the employees of the departments of a given company. Specifically, each
employee has associated a $\attr{Boss}$ who is, in his turn, an employee.

\paragraph{Translating Non-recursive Rules.}$\ $

Non recursive rules are translated in a quite standard way in SQL. The only exceptions are made
for rules containing aggregate functions and rules containing built-ins. The general format of
the SQL statement generated in the translation is:

\begin{quote}

INSERT INTO head($r$) (Translate\_SQL($r$))

\end{quote}

\noindent where {\em head(r)} returns the relation associated with the head of $r$; this task is
carried out by considering the mappings specified in the auxiliary directives. {\em
Translate\_SQL(r)} takes into account the kind of rule (e.g., if it contains negation or
built-ins, etc.) and calls the suitable transformation function. These functions are described
next.

\paragraph{Translating Positive Rules.}$\ $

Intuitively, the SQL statement for positive rules is composed as follows: the SELECT part is
determined by the variable bindings between the head and the body of the rule. The FROM part of
the statement is determined by the predicates composing the body of the rule; variable bindings
between body atoms and constants determine the WHERE conditions of the statement. Finally, an
EXCEPT part is added in order to eliminate tuple duplications. The behaviour of function {\em
TranslatePositiveRule} is well described by the following example.

\begin{example} Consider the following rule:
\[
\rel{q_0}(\attr{Ename})\ \derives
\rel{employee}(\attr{Ename},\attr{100.000},\attr{Dep},\attr{Boss}),
\rel{department}(\attr{Dep},\attr{rossi}).
\]
\noindent which returns all the employees working at the department whose chief is $\attr{rossi}$
and having a yearly salary of $\attr{100.000}$ euros. The corresponding SQL statement is the
following\footnote{Here and in the following we use the notation t.att$_i$ to indicate the i-th
attribute of the table t. Actual attribute names are determined from the auxiliary directives.}:

\begin{quote}
\hspace*{0.0cm}\small{INSERT INTO $q_0$  ( \\
\hspace*{0.5cm}SELECT \ employee.$att_1$ FROM employee, \ department\\
\hspace*{0.5cm}WHERE employee.$att_3$ \ = \ department.$att_1$ AND department.$att_2$='rossi'\\
\hspace*{1.0cm}AND employee.$att_2$=100.000 EXCEPT (SELECT * FROM $q_0$)) } $\punto$
\end{quote}

\end{example}

\paragraph{Translating rules with negated atoms.}$\ $

Intuitively, the construction of the SQL statement for this kind of rule is carried out as
follows: the positive part of the rule is handled in a way very similar to what has been shown
for function {\em TranslatePositiveRule}; then, each negated atom is handled by a corresponding
NOT IN part in the statement. The behaviour of function {\em TranslateRuleWithNegation} is well
illustrated by the following example.

\begin{example}

The following program computes (using the goal $topEmployee$) the employees which have no other
boss than the director.

\[
\begin{array}{lcl}
topEmployee(Ename)&\derives & \rel{employee}(\attr{Ename},\attr{Salary},\attr{Dep},\attr{Boss}),\\
                   &          & \rel{department}(\attr{Dep},\attr{Boss}),\\
                   &        & \naf\ otherBoss(Ename,Boss).\\
otherBoss(Ename,Boss)&\derives & \rel{employee}(\attr{Ename},\attr{Salary},\attr{Dep},\attr{Boss}),\\
                   &            &  \rel{employee}(\attr{Boss},\attr{Salary},\attr{Dep},\attr{Boss1}).\\
\end{array}
\]

\noindent The first rule above is translated to the following SQL statement:

\begin{quote}
\hspace*{0.0cm}\small{INSERT INTO topEmployee ( \\
\hspace*{0.5cm}SELECT employee.att$_1$ FROM employee, department\\
\hspace*{0.5cm}WHERE (employee.att$_3$=department.att$_1$) AND (employee.att$_4$=department.att$_2$) \\
\hspace*{1.0cm}AND (employee.att$_1$, employee.att$_4$)\\
\hspace*{1.0cm}NOT IN (SELECT otherBoss.att$_1$, otherBoss.att$_2$ FROM otherBoss ) \\
\hspace*{0.5cm}EXCEPT (SELECT * FROM topEmployee)) } $\punto$
\end{quote}
\end{example}

\paragraph{Translating rules with built-in predicates.}$\ $

As pointed out in Section \ref{sec:language}, in addition to user-defined predicates some
comparative and arithmetic predicates are provided by the reasoning language. When running a
program containing built-in predicates, the range of admissible values for the corresponding
variables must be fixed. We map this necessity in the working database by adding a restriction
based on the maximum value allowed for variables. Moreover, in order to allow mathematical
operations among attributes, \dlvdb requires the types of attributes to be properly defined in
the auxiliary directives.

The function for translating rules containing built-in predicates is a slight variation of the
function for translating positive rules. As a matter of fact, the presence of a built-in
predicate in the rule implies just adding a corresponding condition in the WHERE part of the
statement. However, if the variables specified in the built-in are not bound to any other
variable of the atoms in the body, a \#maxint value must be exploited to bound that variable to
its admissible range of values.

\begin{example} \label{builtin}
The  program:

\[
\begin{array}{lcl}
q_1(Ename) :- employee(Ename, Salary, Dep, Boss), \ Salary > 100.000.
\end{array}
\]

\noindent is translated to the SQL statement:

\begin{quote}\small{
\hspace*{0.0cm}INSERT INTO $q_1$  \\
\hspace*{0.5cm}(SELECT employee.att$_1$ FROM employee WHERE employee.att$_2$ $>$ 100.000\\
\hspace*{0.5cm} EXCEPT (SELECT * FROM $q_1$))} $\punto$
\end{quote}
\end{example}

\paragraph{Translating rules with aggregate atoms.} $\ $

In Section \ref{sec:language} we introduced the syntax and the semantics of DLP with aggregates.
We have also shown that specific safety conditions must hold for each rule containing aggregate
atoms, in order to guarantee the computability of the corresponding rule. As an example,
aggregate atoms can not contain predicates mutually recursive with the head of the rule  they are
placed in; from our point of view, this implies that the truth values of each aggregate function
can be computed once and for all before evaluating the corresponding rule (which can be, in its
turn, recursive).

Actually, the process that rewrites input programs before their execution, automatically rewrites
each rule containing some aggregate atom in such a way that it follows a standard format (we call
this process standardization in the following). Specifically, given a generic rule of the form:

\[
\begin{array}{lcl}
head & \derives & body,  f(\{Vars:Conj\}) \prec Rg.
\end{array}
\]

\noindent where $Conj$ is a generic conjunction and $Rg$ is a guard, the system automatically
translates this rule to a pair of rules of the form

\[
\begin{array}{lcl}
 auxAtom & \derives & Conj, BindingAtoms. \\
 head & \derives & body, f(\{Vars:auxAtom\}) \prec Rg.
\end{array}
\]

\noindent where $auxAtom$ is a standard rule containing both $Conj$ and the atoms
($BindingAtoms$) necessary for the bindings of $Conj$ with $body$ and/or $head$. Note that
$auxAtom$ contains only those attributes of $Conj$ that are strictly necessary for the
computation of $f$ and, consequently, it may have far less (and can not have more) attributes
than those present in $Conj$.

In our approach we rely on this standardization to translate this kind of rule to SQL; clearly
only the second rule, containing the aggregate function, is handled by the function we are
presenting next; in fact, the first rule is automatically translated by one of the already
presented functions.

Intuitively, the objective of our translation  is to create an SQL view $auxAtom\_supp$ from
$auxAtom$ which contains all the attributes necessary to bind $auxAtom$ with the other atoms of
the original rule and a column storing the results of the computation of $f$ over $auxAtom$; the
original aggregate atom is then replaced by this view and guard conditions are suitably
translated by logic conditions between variables. At this point, the resulting rule is a standard
rule not containing aggregate functions and can be then translated by one of the functions we
have presented previously; clearly enough, in this process, the original input rule $r$ must be
modified to have a proper translation of its ``standard'' part. The function is shown in Figure
\ref{alg:TranslateRuleWithAggregates}; it receives a rule $r$ with aggregates as input and
returns both the SQL views for the aggregate functions in $r$ and the modified (standard) $r$,
which will be handled by standard translation functions\footnote{Here and in the following we use
the operator $+$ to denote the ``append'' operator between strings.}.

\begin{figure}
{\footnotesize
    \begin{tabular}{l}
    \hline
    \hspace*{0.0cm}{Function {\bf TranslateAggregateRule}({\bf VAR} $r$: \DLPA rule): SQL statement} \\
    \hspace*{0.0cm}{\kbegin}  \\
    \hspace*{0.5cm}{\kfor\ each $a$ in aggr\_atom($r$) \kdo\ \kbegin} \\
    \hspace*{1.0cm}{$aux$:=aux\_atom($a$);} \\
    \hspace*{1.0cm}{SQL:=``CREATE VIEW '' + $aux$ +``\_supp'' +} \\
    \hspace*{1.5cm}{``AS (SELECT ''+ bound\_attr($a$) + ``, '' +}\\
    \hspace*{2.0cm}{aggr\_func($a$) + ``('' + aggr\_attr($a$) + ``) '' + } \\
    \hspace*{1.5cm}{``FROM '' + $aux$ + ``GROUP BY '' + bound\_attr($a$) + ``)'';} \\
    \hspace*{1.0cm}{removeFromBody($r$, $a$);} \\
    \hspace*{1.0cm}{addToBody($r$, aux\_atom\_supp($a$));} \\
    \hspace*{1.0cm}{addToBody($r$, guards($a$));} \\
    \hspace*{0.5cm}{\kend;} \\
    \hspace*{0.5cm}{\kreturn\ SQL;} \\
    \hspace*{0.0cm}{\kend .} \\
    \hline
    \end{tabular}
} \hfill
 \caption{Function TranslateAggregateRule} \label{alg:TranslateRuleWithAggregates}
\end{figure}

Here function {\em aggr\_atom($r$)} returns the aggregate atoms present in $r$; {\em
aux\_atom($a$)} returns the auxiliary atom corresponding to $Conj$ of $a$ and automatically
generated by the standardization. Function {\em bound\_attr($a$)} yields in output the attributes
of the atom $a$ bound with attributes of the other atoms in the rule, whereas {\em
aggr\_attr($a$)} returns the attribute which the aggregation must be carried out onto (the first
variable in $Vars$). {\em aggr\_func($a$)} returns the SQL aggregation statement corresponding to
the aggregate function of $a$. Functions {\em removeFromBody} and {\em addToBody} are responsible
of altering the original rule $r$ to make it standard (without aggregates). In particular, {\em
removeFromBody($r$, $a$)} removes the aggregate atom $a$ from the rule $r$, whereas {\em
addToBody} adds both {\em aux\_atom\_supp($a$)} and {\em guards($a$)} to $r$. Note that {\em
aux\_atom\_supp($a$)} yields in output the name of the atom corresponding to the just created
auxiliary view, whereas {\em guards($a$)} converts the guard of the aggregate atom $a$ in a logic
statement between attributes in the rule.

\begin{example}

Consider the following rule computing the departments which spend for the salaries of their
employees, an amount greater than a certain threshold, say 100000:

\begin{quote}
$\rel{costlyDep}(\attr{Dep}) \derives\ \rel{department}(\attr{Dep},\_\ ),$

$\hspace*{2.5cm}\#sum\{\attr{Salary},\attr{Ename}:
\rel{employee}(\attr{Ename},\attr{Salary},\attr{Dep},\_\ )\} > 100000.\footnote{Note that Ename
is needed to sum also the salaries of employees earning the same amount (see the discussion on
sets/multisets in \cite{dell-etal-2003a}).} $
\end{quote}

\noindent The standardization automatically rewrites this rule as:

\begin{quote}
$\rel{aux\_emp}(\attr{Salary},\attr{Ename},\attr{Dep})  \derives\  \rel{department}(\attr{Dep},
\_\ ),$

$\hspace*{4.7cm}\rel{employee}(\attr{Ename},\attr{Salary},\attr{Dep},\_\ ).$

$\rel{costlyDep}(\attr{Dep}) \derives\ \rel{department}(\attr{Dep},\_\ ),$

$\hspace*{2.5cm}\#sum\{\attr{Salary},\attr{Ename}:
\rel{aux\_emp}(\attr{Salary},\attr{Ename},\attr{Dep})\} > 100000. $

\end{quote}

\noindent The first rule is treated as a standard positive rule and is translated to:

\begin{quote}
\small{
\hspace*{0.0cm}INSERT INTO aux\_emp ( \\
\hspace*{0.5cm}SELECT employee.att$_2$, employee.att$_1$, department.att$_1$\\
\hspace*{0.5cm}FROM department, employee WHERE department.att$_1$ = employee.att$_3$ \\
\hspace*{0.5cm}EXCEPT (SELECT * FROM aux\_emp)) }
\end{quote}

\noindent The second rule is translated to:

\begin{quote}
\small{
\hspace*{0.0cm}CREATE VIEW aux\_emp\_supp AS ( \\
\hspace*{0.5cm}SELECT aux\_emp.$att_3$, SUM (aux\_emp.$att_1$) FROM aux\_emp\\
\hspace*{0.5cm}GROUP BY aux\_emp.$att_3$) \\

\hspace*{0.0cm}INSERT INTO costlyDep ( \\
\hspace*{0.5cm}SELECT department.$att_1$ FROM department, aux\_emp\_supp\\
\hspace*{0.5cm}WHERE department.$att_1$ = aux\_emp\_supp.$att_1$ AND aux\_emp\_supp.$att_2 > 100000 $\\
\hspace*{0.5cm}EXCEPT (SELECT * FROM costlyDep)) } $\punto$
\end{quote}

\end{example}

\paragraph{Translating recursive rules.}$\ $

As previously pointed out, our program evaluation strategy exploits a refined version of the
Semi-Naive method. This is based on the translation of a recursive rule into a  non recursive SQL
statement operating alternatively on  standard and differential versions of the relations
associated with recursive predicates. Each time this statement is executed by the algorithm, it
must compute just the new values for the predicate in the head that can be obtained from the
values computed in the last two iterations of the fixpoint.

Intuitively, the translation algorithm must first select the proper combinations of standard and
differential relations from the rule $r$ under consideration; then, for each of these
combinations, it must rewrite $r$ in a corresponding rule $r'$. Each $r'$ thus obtained is non
recursive and, consequently, it can be handled by Function TranslateNonRecursiveRule. Algorithm
TranslateRecursiveRule is shown in Figure \ref{alg:TranslateRecursiveRule}.

Here, functions {\em TranslateAggregateRule} and {\em TranslateNonRecursiveRule} have been
introduced previously. Function {\em hasAggregate($r$)} returns true if $r$ contains aggregate
functions. Function {\em RecursivePredicates($r$)} returns the number of occurrences of recursive
predicates in the body of $r$; {\em $\Delta$head($r$)} returns the differential version of the
relation corresponding to the head of $r$. Function {\em setHead($r'$, $p$)} sets the head of the
rule $r'$ to the predicate $p$; analogously, function {\em addToBody($r'$, $p$)} adds to the body
of $r'$ a conjunction with the predicate $p$. Function {\em bit(j,i)} returns the $j$-th bit of
the binary representation of $i$.

It is worth noticing that the execution of the queries resulting from function
TranslateRecursiveRule implement  function $\mathit{EVAL\_DIFF}$ for $r$ (see the algorithm of
Figure \ref{alg:semi-naive-mod}).

\begin{figure}

{\footnotesize
    \begin{tabular}{l}
    \hline
    \hspace*{0.0cm}{Function {\bf TranslateRecursiveRule}($r$: \DLPA rule ): $SQL$ statement} \\
    \hspace*{0.0cm}{\kbegin}  \\
    \hspace*{0.5cm}{$SQL$:="";} \\
    \hspace*{0.5cm}{\kif (hasAggregate($r$)) \then} \\
    \hspace*{1.0cm}{$SQL$:=TranslateAggregateRule($r$);} \\
    \hspace*{0.5cm}{n:=$2^{RecursivePredicates(r)}$-1} \\
    \hspace*{0.5cm}{$SQL$:=$SQL$+"INSERT INTO " + $\Delta$head($r$) + "(";} \\
    \hspace*{0.5cm}{\kfor \ i:=1 to n \kdo \ \kbegin} \\
    \hspace*{1.0cm}{Let $r'$ be a rule;} \\
    \hspace*{1.0cm}{setHead($r'$, $\Delta$head(r));} \\
    \hspace*{1.0cm}{\kfor \ each non recursive predicate $q_j$ in body(r) \kdo} \\
    \hspace*{1.5cm}{addToBody($r'$, $q_{j}$);} \\
    \hspace*{1.0cm}{\kfor \ each recursive predicate $p_j$ in body(r) \kdo} \\
    \hspace*{1.5cm}{\kif \ (bit(j,i)=0) \kthen \ addToBody($r'$, $p_{j}^{k-2}$);} \\
    \hspace*{1.5cm}{\kelse \ addToBody($r'$, $\Delta p_{j}^{k-1}$);} \\
    \hspace*{1.0cm}{\kif \ (i $\neq$ 1) \ SQL:=SQL+"UNION ";} \\
    \hspace*{1.0cm}{SQL:=SQL + TranslateNonRecursiveRule($r'$);} \\
    \hspace*{0.5cm}{\kend;} \\
    \hspace*{0.5cm}{$SQL$:=$SQL$ + ")";} \\
    \hspace*{0.5cm}{\kreturn\ $SQL$;} \\
    \hspace*{0.0cm}{\kend .} \\
    \hline
    \end{tabular}
} \hfill
 \caption{Function TranslateRecursiveRule} \label{alg:TranslateRecursiveRule}
\end{figure}

\begin{example} Consider the situation in which we need to know
whether the employee $e_1$ is the boss of the employee $e_n$ either directly or by means of a
number of employees $e_2,..,e_n$ such that $e_1$ is the boss of $e_2$, $e_2$ is the boss of
$e_3$, etc. Then, we have to evaluate the program:
\[
\begin{array}{llcl}
r_1: &  q_2(E_1,E_2)\ & \derives\ & \rel{employee}(\attr{E_1},\attr{Salary},\attr{Dep},\attr{E_2}).\\
r_2: & q_2(E_1,E_3)\ & \derives\ & q_2(E_1,E_2),\ q_2(E_2,E_3).
\end{array}
\]

\noindent containing the recursive rule $r_2$. This program cannot be evaluated in one single
iteration of the Semi-Naive computation. Rule $r_1$ is not recursive; it is translated by
Function {\em TranslatePositiveRule} to the following SQL which is evaluated once:

\begin{quote}
\hspace*{0.0cm}\small{INSERT INTO $q_{2}$ ( SELECT $employee.att_1$, $employee.att_4$ FROM $employee$ \\
\hspace*{0.5cm}EXCEPT (SELECT * FROM $q_{2}$))}
\end{quote}

Rule $r_2$ is first translated by Function {\em TranslateRecursiveRule} to the temporary set of
rules:

\[
\begin{array}{llcl}
r_2^{'}: & \Delta q_{2}^{k}(E_1,E_3)\ & \derives\ & q_{2}^{k-2}(E_1,E_2),\ \Delta q_{2}^{k-1}(E_2,E_3). \\
 & \Delta q_{2}^{k}(E_1,E_3)\ & \derives\ & \Delta q_{2}^{k-1}(E_1,E_2),\ q_{2}^{k-2}(E_2,E_3). \\
 & \Delta q_{2}^{k}(E_1,E_3)\ & \derives\ & \Delta q_{2}^{k-1}(E_1,E_2),\ \Delta
q_{2}^{k-1}(E_2,E_3).
\end{array}
\]

\noindent which is translated to:

\begin{quote}
\hspace*{0.0cm}\small{INSERT INTO $\Delta q_{2}^{k}$ ( \\
\hspace*{0.5cm}SELECT $q_{2}^{k-2}.att_1$, $\Delta q_{2}^{k-1}.att2$ FROM $q_{2}^{k-2}$,$\Delta q_{2}^{k-1}$  WHERE ($q_{2}^{k-2}$.$att_2$=$\Delta q_{2}^{k-1}.att_1$)\\
\hspace*{0.5cm}EXCEPT (SELECT * FROM $\Delta q_{2}^{k}$) \\
\hspace*{0.5cm}UNION \\
\hspace*{0.5cm}SELECT $\Delta q_{2}^{k-1}.att_1$, $q_{2}^{k-2}.att2$ FROM $\Delta q_{2}^{k-1}$, $q_{2}^{k-2}$ WHERE ($\Delta q_{2}^{k-1}.att_2$=$q_{2}^{k-2}$.$att_1$)\\
\hspace*{0.5cm}EXCEPT (SELECT * FROM $\Delta q_{2}^{k}$) \\
\hspace*{0.5cm}UNION \\
\hspace*{0.5cm}SELECT $\Delta q_{2}^{k-1}.att_1$, $\Delta q_{2}^{k-1}\_1.att2$ FROM $\Delta q_{2}^{k-1}$, $\Delta q_{2}^{k-1}$ AS $\Delta q_{2}^{k-1}$\_1\\
\hspace*{0.5cm}WHERE ($\Delta q_{2}^{k-1}$.$att_2$=$\Delta q_{2}^{k-1}\_1.att_1$) \\
\hspace*{0.5cm}EXCEPT (SELECT * FROM $\Delta q_{2}^{k}$))
 }
\end{quote}

\noindent Actually, the real implementation of this function adds, for performance reasons, also
the following parts to the statement above:

\begin{quote}
\small{ \noindent
 \hspace*{0.5cm}EXCEPT (SELECT * FROM $\Delta q_{2}^{k-1}$) \\
 \hspace*{0.5cm}EXCEPT (SELECT * FROM $q_{2}^{k-2}$) }
\end{quote}

\noindent Note that, following Algorithm {\em Differential Semi-Naive} (Figure
\ref{alg:semi-naive-mod}), $q_{2}^{k-2}$ and $\Delta q_{2}^{k-1}$ are first initialized with the
result of the evaluation of $r_1$ (stored in $q_{2}$ - see instructions (2) and (3) in Figure
\ref{alg:semi-naive-mod}). Then, the SQL above is iteratively executed until the fixpoint is
reached. Note that, the aforementioned process executes instructions (1)-(5) of the algorithm in
Figure \ref{alg:semi-naive-mod}. The update of $q_{2}^{k-2}$ and $\Delta q_{2}^{k-1}$ from one
iteration to the subsequent one is carried out by instructions (6) and (7) in a straightforward
way.
 $\punto$
\end{example}

\subsubsection{A complete example}

\begin{example} \label{example3}

Consider the datalog program presented in Example \ref{example} and the mappings shown in Figure
\ref{fig:mappings}. The complete query plan derived by \dlvdb for them is:

\vspace*{-0.0cm} {\small
\begin{tabular}{ll}
(1)&INSERT INTO destinations\_rel\\
 & (SELECT f.FromX, f.ToY, f.Company FROM flight\_rel AS f)\\
&\\
(2) & INSERT INTO destinations\_rel \\
 & (SELECT f.FromX, f.ToY, c.Company2 FROM flight\_rel AS f, codeshare\_rel AS c\\
& WHERE (f.Id=c.FlightId) AND (f.Company=c.Company1) \\
& EXCEPT (SELECT * FROM destinations\_rel))
\end{tabular}}
{\small
\begin{tabular}{ll}
(3)& INSERT INTO d\_destinations\_rel\\
& (SELECT d1.FromX, d2.ToY, d1.Company \\
& FROM d1\_destinations\_rel AS d1, destinations\_rel AS d2\\
& WHERE (d1.ToY=d2.FromX) AND (d1.Company=d2.Company) \\
& UNION \\
& SELECT d1.FromX, d2.ToY, d1.Company \\
 & FROM destinations\_rel AS d1, d1\_destinations\_rel AS d2\\
& WHERE (d1.ToY=d2.FromX) AND (d1.Company=d2.Company)\\
& UNION\\
& SELECT d1.FromX, d2.ToY, d1.Company \\
& FROM d1\_destinations\_rel AS d1, d1\_destinations\_rel AS d2\\
& WHERE (d1.ToY=d2.FromX)AND (d1.Company=d2.Company)\\
& EXCEPT (SELECT * FROM d1\_destinations\_rel)\\
& EXCEPT (SELECT * FROM destinations\_rel)\\
& EXCEPT (SELECT * FROM d\_destinations\_rel))
\end{tabular}
}  \vspace*{-0.0cm}

\noindent SQL statements (1) and (2) are executed only once, since they correspond to non
recursive rules. On the contrary, the statement (3) is executed several times, until the least
fixpoint is reached, i.e. {\tt d\_destinations\_rel} is empty. Note that {\tt
d\_destinations\_rel} and {\tt d1\_destinations\_rel} correspond, respectively, to $\Delta
head(r)$ and $\Delta p^{k-1}$ introduced in function TranslateRecursiveRule; as shown in Section
\ref{sub:evaluation} the evaluation algorithm suitably updates the tuples of {\tt
destinations\_rel} from the new values derived at each iteration in {\tt d\_destinations\_rel}.
$\punto$
\end{example}

\section{System Architecture}
\label{sec:Architecture}

In this section we present the general architecture of our system. It has been designed as an
extension of the \dlv system \cite{leon-etal-2002-dlv}, which allows both the evaluation of logic
programs directly on databases and the handling of input and output data distributed on several
databases. It combines the expressive power of \dlv (and the optimization strategies implemented
in it) with the efficient data management features of DBMSs \cite{ullm-etal-00}.

As previously pointed out, the system provides two, quite distinct, functioning modalities,
namely the direct database execution and the main-memory execution modality. In the following we
present the two corresponding architectures separately.

\subsection{Architecture of the direct database execution (DLV$^{DB}$) } \label{sub:dlvdbArchitecture}

\begin{figure}[t]

\centerline{\psfig{figure=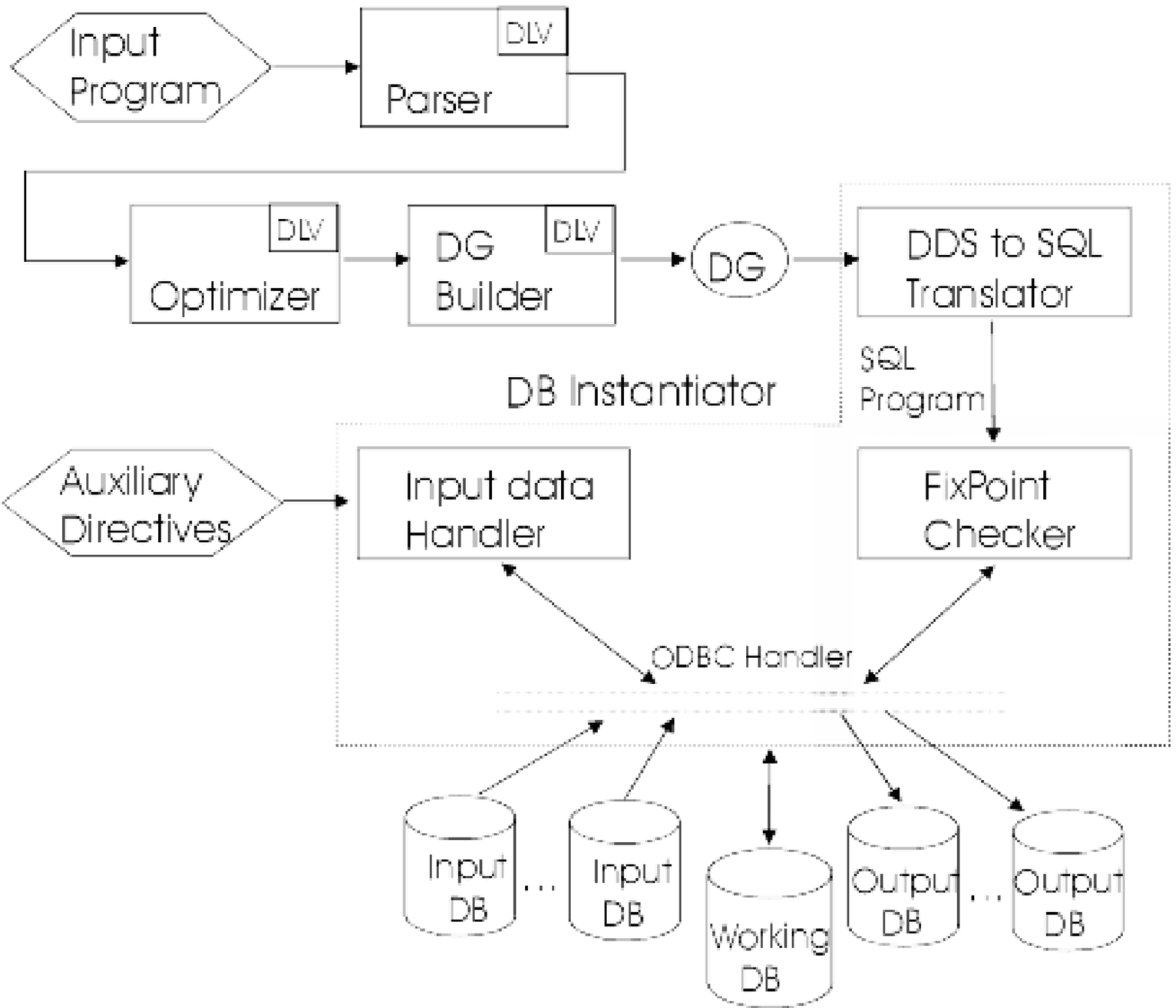,width=8cm,height=7cm}} \caption{Architecture of
\dlvdb.} \label{ArchitectureFigure}
\end{figure}

Figure \ref{ArchitectureFigure} illustrates the architecture of the system for the direct
database execution. In the figure, the boxes marked with DLV are the ones already developed in
the DLV system. An input program $\cal P$ is first analyzed by the Parser which encodes the rules
in the intensional database (IDB) in a suitable way and builds an extensional database (EDB) in
main-memory data structures from the facts specified directly in the program (if any). As for
facts already stored in database relations, no EDB is produced in main-memory. After this, the
Optimizer applies a rewriting procedure in order to get a program $\cal P'$, equivalent to $\cal
P$, that can be evaluated more efficiently; some of the operations carried out by this module
have been highlighted in Section \ref{sub:evaluation}. The Dependency Graph Builder computes the
dependency graph of $\cal P'$, its connected components and a topological ordering of these
components. Finally, the DB Instantiator module, the core of the system, is activated.

The DB Instantiator module receives: {\em (i)} the IDB and the EDB (if not empty) generated by
the parser, {\em (ii)} the Dependency Graph (DG) generated by the dependency graph builder and
{\em (iii)} the auxiliary directives specifying the needed interactions between \dlvdb and the
databases. It evaluates the input program through the bottom-up fixpoint evaluation strategy
shown in Section \ref{sec:Implementation}. Since the input program is supposed to be normal and
stratified (see Section \ref{sec:language}), the DB Instantiator evaluates completely the program
and no further modules must be employed after it.

All the instantiation steps are performed directly on the working database through the execution
of SQL statements and no data is loaded in main-memory from the databases in any phase of the
process. This allows \dlvdb to be completely independent of the dimension of both the input data
and the number of facts generated during the evaluation.

Communication with databases is performed via ODBC. This allows \dlvdb both to be independent
from a particular DBMS  and to handle databases distributed over the Internet.

It is important to point out that the architecture of \dlvdb has been designed in such a way to
fully exploit  optimizations both from logic theory and from database theory. In fact, the
actually evaluated program is the one resulting from the Optimizer module which applies program
rewriting techniques aiming to simplify the evaluation process and to reduce the dimension of the
involved relations (see Section \ref{sub:evaluation}). Then, the execution of the SQL statements
in the query plan exploit data-oriented optimizations implemented in the DBMS. As far as this
latter point is concerned, we have experienced that the kind of DBMS handling the working
database for \dlvdb may significantly affect system performance; in fact, when \dlvdb was coupled
with highly sophisticated DBMSs it generally showed better performance in handling large amounts
of data w.r.t. the same executions when coupled with less sophisticated DBMSs.

The observation above points out both the importance of data-oriented optimizations and a
potential advantage of \dlvdb w.r.t. deductive systems operating on ad-hoc DBMSs. In fact, \dlvdb
can be easily coupled with the most efficient DBMS available at the time being used (provided
that it supports standard SQL), whereas the improvement of an ad-hoc DBMS is a more difficult
task.

\subsection{Architecture of the main-memory execution (DLV$^{IO}$)}

\begin{figure}
\begin{center}
\epsfig{file=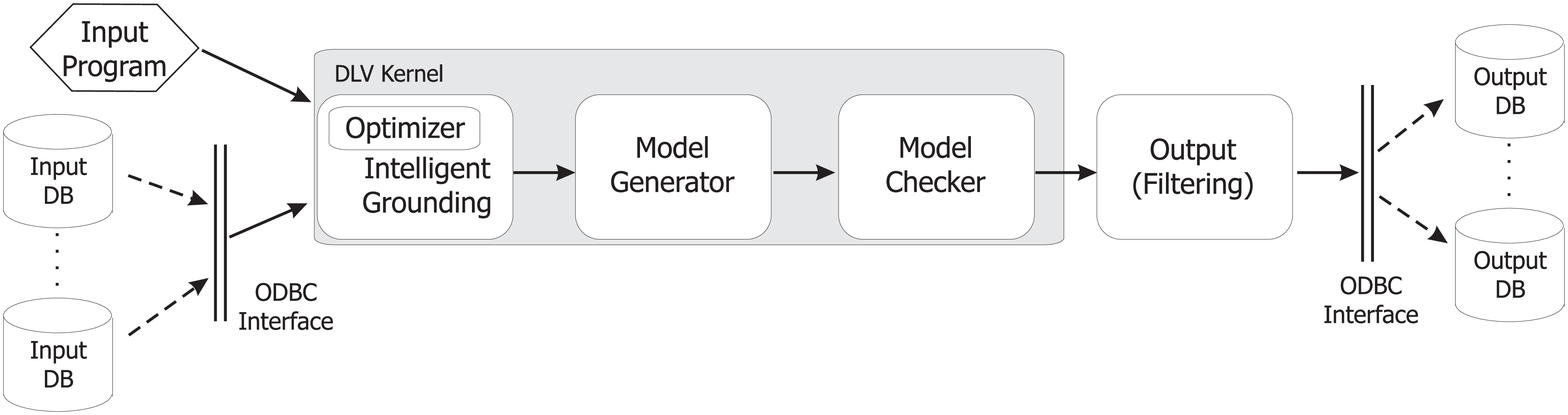,height=3.0cm}
\end{center}
 \caption{Architecture of \dlvio} \label{fig:architecture-io}
\end{figure}

The architecture of \dlvio is illustrated in Figure \ref{fig:architecture-io}. It extends the
classical \dlv architecture with ODBC functionalities to import/export data from/to database
relations. The main-memory execution modality acts just as an interface (based on ODBC
connections) between the external databases and the standard \dlv program.

In more detail, input data can be supplied both by regular files and by relational tables
accessed via ODBC as specified by the \#import commands. Specifically, for each \#import command
the system retrieves data from the corresponding table ``row by row'' through the SQL query
specified by the user and creates one atom in main-memory (in the format required by \dlv) for
each selected tuple. The name of each imported atom is set to {\em predname}, and is considered
as a fact. Possible facts residing in text files are fed into \dlv in the standard way. All the
data is fetched in  main-memory before any evaluation task is carried out.

The \dlv kernel (the shaded part in the figure) then produces answer sets one at a time. It
consists of three major components: the ``Intelligent Grounding''\footnote{It incorporates the
Parser, the Optimizer and the DG Builder depicted in Figure \ref{ArchitectureFigure}.}, the
``Model Generator'', and the ``Model Checker'' modules; these share a main data structure, the
``Ground Program''. It is created by the Intelligent Grounding using differential (and other
advanced) database techniques together with suitable main-memory data structures, and used by the
Model Generator and the Model Checker. The Ground Program is guaranteed to have exactly the same
answer sets as the original program. For some syntactically restricted classes of programs (e.g.\
stratified programs), the Intelligent Grounding module already computes the corresponding answer
sets.

For harder problems, most of the computation is performed by the Model Generator and the Model
Checker.  Roughly, the former produces some ``candidate'' answer sets (models)
\cite{fabe-etal-99b,fabe-etal-2001a}, the stability and minimality of which are subsequently
verified by the latter.

The Model Checker  verifies whether the model at hand is an answer set.  This task is very hard
in general, because checking the stability of a model is known to be co-\NP-complete. However,
this module exploits the fact that minimal model checking --- the hardest part --- can be
efficiently performed for the relevant class of \emph{head-cycle-free} (HCF) programs
\cite{bene-dech-94,bene-dech-96}.

Each time an answer set $M$ is found, ``Filtering'' is invoked, which performs some
post-processing, controls continuation or abortion of the computation, and possibly stores the
output data in the corresponding relational tables as specified by the \#export commands. In
particular, if an \#export command from {\em predname} to {\em tablename} is present, the module
generates a new tuple in {\em tablename} for each atom in $M$ having name {\em
predname}\footnote{As previously pointed out, the presence of an \#export command automatically
limits the system to generate the first answer set only.}.

\section{Experiments and Benchmarks} \label{sec:experiments}

In this section we present our experimental framework and the results obtained  comparing the
\dlvdb system with several state-of-the-art systems. Benchmarks have been designed following the
guidelines, problems and data structures proposed in \cite{ban-ram-88} and \cite{greco-03} to
assess the performance of deductive database systems. Roughly speaking, problems used in
\cite{ban-ram-88} and \cite{greco-03} basically resort to the execution of some recursive queries
on a variety of data structures. The main goal of our experiments was to evaluate the deductive
capabilities of tested systems for both query answering time and amount of manageable data,
especially with respect to the direct database execution of our system.

All tests have been carried out on a Pentium 4 machine with a 1.4 GHz CPU and 512 Mbytes of RAM.

\subsection{Overview of Compared Systems}
\label{sub:systemsoverview}

In order to provide a comparative and comprehensive analysis with the state-of-the-art systems in
the considered research area, we have compared our system performance, under both execution
modalities (i.e, both \dlvdb and \dlvio), with: {\em (i)} LDL++, because it is one of the most
robust implementations of deductive database systems; {\em (ii)} XSB, as an efficient
implementation of the Top-Down evaluation strategy; {\em (iii)} Smodels, one of the most widely
used Answer Set Programming systems together with \dlv; {\em (iv)} three commercial DBMSs
supporting the execution of recursive queries. Note that the licence of use of such DBMSs does
not allow us to explicitly mention them in the paper; as a consequence, in the following, we call
them simply DB-A, DB-B and DB-C. The reader should just know they are the three top-level
commercial database systems currently available, which also support recursive queries.

Note that important DBMSs, such as Postgres and MySQL could not be tested; in fact, they do not
support recursive queries, which are the basis for our testing framework. Moreover, as we pointed
out in the Introduction, other logic-based systems such as ASSAT, Cmodels, and CLASP have not
been tested since they use the same grounding layer of Smodels (LParse) and, as it will be clear
in the following, the benchmark programs are completely solved by this layer.

In the following we briefly overview the main characteristics of the tested systems, focusing on
their support to the language and technological capabilities addressed in this work.
Specifically, we consider, for each database system, its capability to express recursive queries
and, for each logic-based system, the expressiveness of its language and its capability to
interact with external DBMSs.

For each system, we used the latest release available at the time tests have been carried out.

\subsubsection{Database systems}

As far as database systems are concerned, it is worth pointing out that none of the considered
ones fully adopt the SQL99 standard for the definition of recursive queries, but proprietary
constructs are introduced by each of them.

In particular, both DB-A and DB-B support the standard recursive functionalities that are needed
for our benchmarks, even if proprietary constructs must be added to the standard SQL99 statement
to guarantee the termination of some kinds of queries. On the contrary, DB-C implements a large
subset of SQL99 features and supports recursion but, as far as recursive queries are concerned,
it exploits proprietary constructs which do not follow the standard SQL99 notation, and whose
expressiveness is lower than that of SQL99; as an example, it is not possible to express unbound
queries within recursive statements (e.g., \emph {all} the pairs of nodes linked by at least one
path in a graph).

\subsubsection{LDL++}
The LDL++ system \cite{arni-etal-03} integrates rule-based programming with efficient secondary
memory access, transaction management recovery and integrity control. The underlying database
engine has been developed specifically within the LDL project and is designed as a virtual-memory
record manager, which is optimized for the situation where the pages containing frequently used
data can reside in main-memory. LDL++ can also be interfaced with external DBMSs, but it is
necessary to implement vendor-specific drivers to handle data conversion and local SQL dialects
\cite{arni-etal-03}. The LDL++ language supports complex terms within facts and rules, stratified
negation, and don't care non-determinism based on stable model semantics. Moreover, LDL++
supports updates through special rules.

In our tests we used version 5.3 of LDL++. Test data have been fed to the system by text files
storing input facts.

\subsubsection{XSB}
The XSB system \cite{Rao*97} is an in­memory deductive database engine based on a Prolog/SLD
resolution strategy called SLG. It supports explicitly locally stratified programs. The inference
engine, which is called SLG-WAM, consists of an efficient tabling engine for definite logic
programs, which is extended by mechanisms for handling cycles through negation. These mechanisms
are negative loop detection, delay and simplification. They serve for detecting, breaking and
resolving cycles through negation.

XSB allows the exploitation of data residing in external databases, but reasoning on such data is
carried out in main-memory. The version of XSB we used in our tests is 2.2.

\subsubsection{SModels}
The SModels system \cite{niem-etal-2000a,niem-simo-97} implements the answer set semantics for
normal logic programs extended by built-in functions as well as cardinality and weight
constraints for domain-restricted programs.

The SModels system takes as input logic program rules in Prolog style syntax. However, in order
to support efficient implementation techniques and extensions, the programs are required to be
{\em domain-restricted} where the idea is the following: the predicate symbols in the program are
divided into two classes, {\em domain predicates} and {\em non-domain} predicates. Domain
predicates are predicates that are defined non-recursively. The main intuition of domain
predicates is that they are used to define the set of terms over which the variable range in each
rule of a program $P$. All rules of $P$ have to be domain-restricted in the sense that every
variable in a rule must appear in a domain predicate which appears positively in the rule body.
In addition to normal logic program rules, SMODELS supports rules with cardinality and weight
constraints, which are similar to $\countagg$ and $\sumagg$ aggregates of \dlv.

SModels does not allow to handle data residing in database relations; moreover, all the stages of
the computation are carried out in main-memory. Finally, it does not support optimization
strategies for bound queries; consequently, the time it needs for executing the same query either
with all parameters unbound or with some parameters bound is exactly the same.

In our tests we used SModels ver. 2.28 with Lparse ver. 1.0.17. Test data have been fed to the
system by text files storing input facts.

\subsubsection{\dlvdb}
It is the direct database execution of our system; in our tests we used a commercial database as
DBMS for the working database. However, to guarantee fairness with the other systems, we did not
set any additional index or key information for the involved relations. We point out again that
any DBMS supporting ODBC could be easily coupled with \dlvdb.

\subsubsection{\dlvio}
It is the main-memory execution modality of the system presented in this paper. Recall that it
basically corresponds to the execution of the standard \dlv system with data loaded from
databases.

\subsection{Benchmark Problems}
\label{sub:problems}

To asses the performance of the systems described above, we carried out several tests using
classical benchmark problems from the context of deductive databases \cite{ban-ram-88,greco-03},
namely \emph{Reachability} and \emph{Same Generation}. The former allows the analysis of basic
recursion capabilities of the various systems on several data structures, whereas the latter
implements a more complex problem and, consequently, allows the capability of the considered
systems to carry out more refined reasoning tasks to be tested.

For each problem, we measured the performance of the various systems in computing three kinds of
queries, namely: unbound queries (identified by the symbol \emph{\q$_{0}$} in the following);
queries with one bound parameter   (\emph{\q$_{1}$}); queries with all bound parameters
(\emph{\q$_{2}$}). Considering these three cases is important because DBMSs and Deductive
Databases generally benefit from query bindings (by ``pushing down'' selections through
relational algebra optimizations, magic set techniques, or, for XSB, top down evaluation),
whereas ASP systems are generally more effective with unbound queries (since they usually compute
the entire models anyway); as a consequence, it is interesting to test all these systems in both
their favorable and unfavorable contexts. It is worth pointing out that some of the tested
systems implement optimization strategies `a la magic set'
\cite{Bancilhon*86,beer-rama-91,Mumick*96,Ross90} (e.g., \dlvdb and LDL++), typical of deductive
databases, or other program rewriting techniques; as a consequence, the actually evaluated
programs are the optimized ones automatically derived by these systems, but the cost of these
rewritings has been always considered in the measure of systems' performance.

In what follows we briefly introduce the two considered problems; the interested reader can find
all details about them in \cite{ban-ram-88}.

\subsubsection{Reachability} Given a directed graph $G=(V,E)$ the solution to the reachability
problem \code{reach}- \code{able(a,b)} determines whether a node $b \in V$ is reachable from a
node $a \in V$ through a sequence of edges in $E$. The input is provided by a relation
\code{edge(X,Y)} where a fact \code{edge(a,b)} states that \code{b} is directly reachable by an
edge from \code{a}.

In database terms, determining all pairs of reachable nodes in $G$ amounts to computing the
transitive closure of the relation storing the edges.

\subsubsection{Same Generation}
Given a parent-child relationship (a tree), the Same Generation problem aims to find pairs of
persons belonging to the same generation. Two persons belong to the same generation either if
they are siblings, or if they are children of two persons of the same generation.

The input is provided by a relation \code{parent(X,Y)} where a fact \code{parent(thomas,}
\code{moritz)} means that \code{thomas} is the parent of \code{moritz}.

\subsection{Benchmark Data Sets}
\label{sub:BenchmarkData}

For each considered problem we exploited several sets of benchmark data structures. For each data
structure various instances of increasing dimensions have been constructed; the size of each
instance is measured in terms of the number of input facts describing it.

\subsubsection{Reachability}
As for the Reachability problem, we considered: {\em (i)} full binary trees, {\em (ii)} acyclic
graphs (a-graphs in the following), {\em (iii)} cyclic graphs (c-graphs in the following), and
{\em (iv)} cylinders \cite{ban-ram-88}.

The \emph{density} $\delta$ of a graph can be measured as $\delta=\frac{\mbox{\# of arcs in the
graph}}{\mbox{\# of possible arcs}}$. We generated various typologies of graph instances,
characterized by values of $\delta$ equal to $0.20$, $0.50$ and $0.75$ respectively. Due to space
constraints, in this paper we report just the results obtained for $\delta=0.20$.

\begin{figure}[t]
\centerline{\psfig{figure=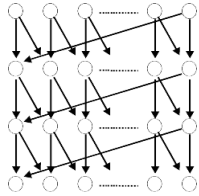,width=2.5cm,height=2.5cm}} \caption{Example of a cylinder
graph.} \label{fig:cylinder}
\end{figure}

Cylinders are particular kinds of acyclic graphs which can be layered; each layer has the same
number of nodes. Each node of the first layer has two outgoing arcs and no incoming arcs, whereas
each node of the last layer has two incoming arcs and no outgoing arcs; finally, each node of an
internal layer has two incoming and two outgoing arcs. An example of a cylinder is shown in
Figure \ref{fig:cylinder}. A cylinder has then a width and a height; as a consequence, the ratio
$\rho=\frac{\mbox{width}}{\mbox{height}}$ can be exploited to characterize a cylinder. We
generated various categories of cylinders having $\rho$ equal to $0.5$, $1.0$ and $1.5$
respectively. Due to space constraints, in this paper we report just the results obtained for
$\rho=1$.

Graphs have been generated using the Stanford GraphBase \cite{knut-94} library whereas trees and
cylinders have been generated using ad-hoc procedures, since they are characterized by a regular
structure.

\subsubsection{Same Generation}
As far as the Same Generation problem is concerned, we exploited full binary trees as input data
structures.

\subsection{Problem Encodings}
\label{sub:Encoding}

We have used general encodings for the two considered problems in a way which tests the various
systems under generic conditions; specifically, we used ``uniform'' queries, i.e. queries whose
structure must not be modified depending on the quantity and positions of bound parameters.
Several alternative encodings could have been possible for the various problems, depending also
on the underlying data structures; however, since many other problems of practical relevance can
be brought back to the ones we considered, we preferred to exploit those encodings applicable to
the widest variety of applications.

Due to space constraints we can not list here the encodings exploited in our tests. The
interested reader can find them at the address {\tt http://www.mat.unical.it/}
{terracina/tplp-dlvdb/encodings.pdf}.

Note that, since DB-C does not support the standard SQL99 language, but only a simplified form of
recursion, we have not tested this system along with the other ones. We will discuss encodings
and results obtained for DB-C in a separate section.

\subsection{Results and Discussion}
\label{sub:results}

In our tests we measured the time required by each system to answer the various queries. We fixed
a maximum running time of 12000 seconds (about 3 hours) for each test. In the following figures,
the line of a system stops whenever some query was not solved within this time limit (note that
graphs have a logarithmic scale on the vertical axis).

\begin{figure}
\begin{tabular}{cc}
\small{samegen(X,Y)} & \small{samegen(b1,Y)} \\
\psfig{figure=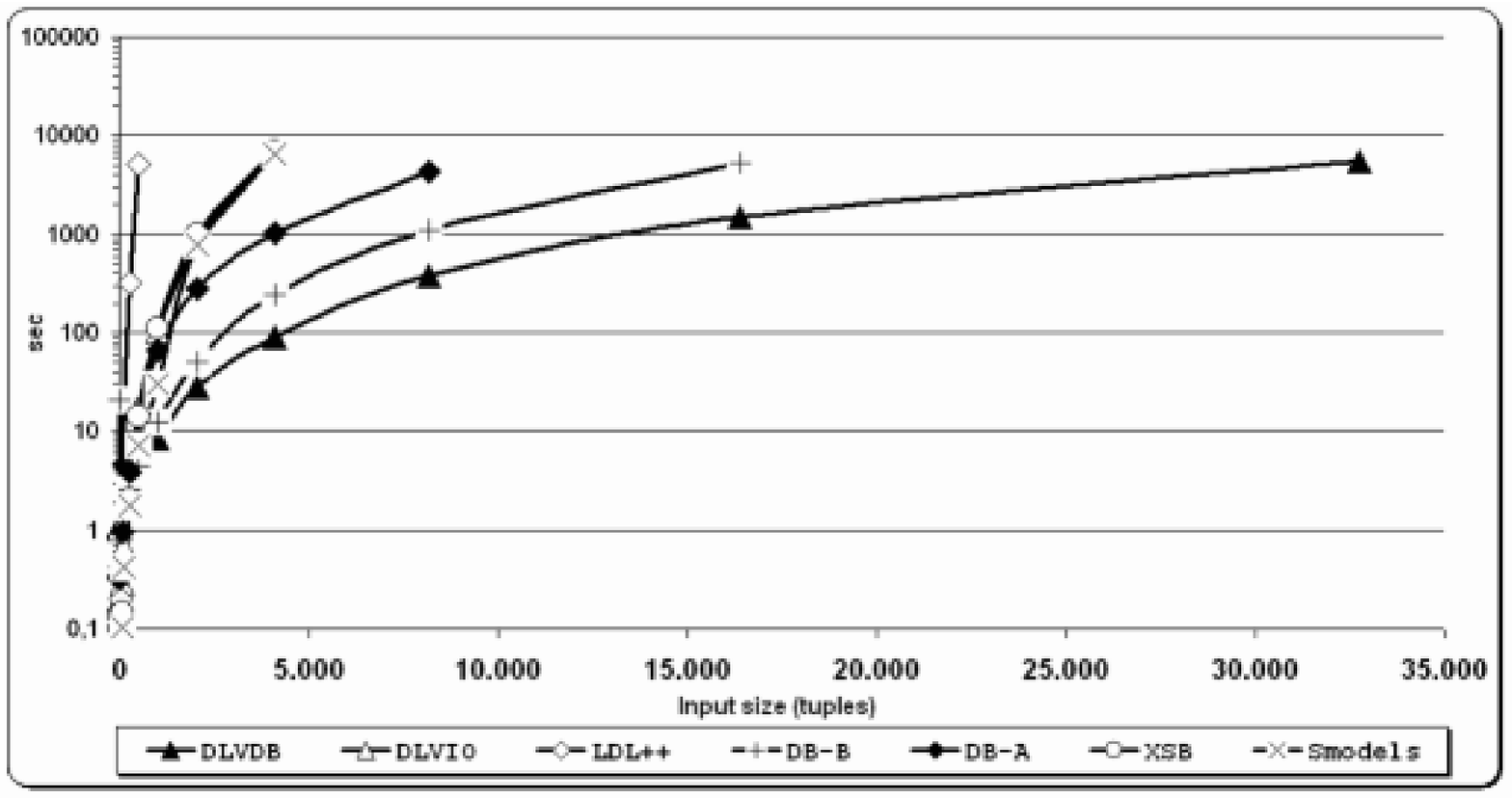,width=6.0cm,height=4.0cm} &
\psfig{figure=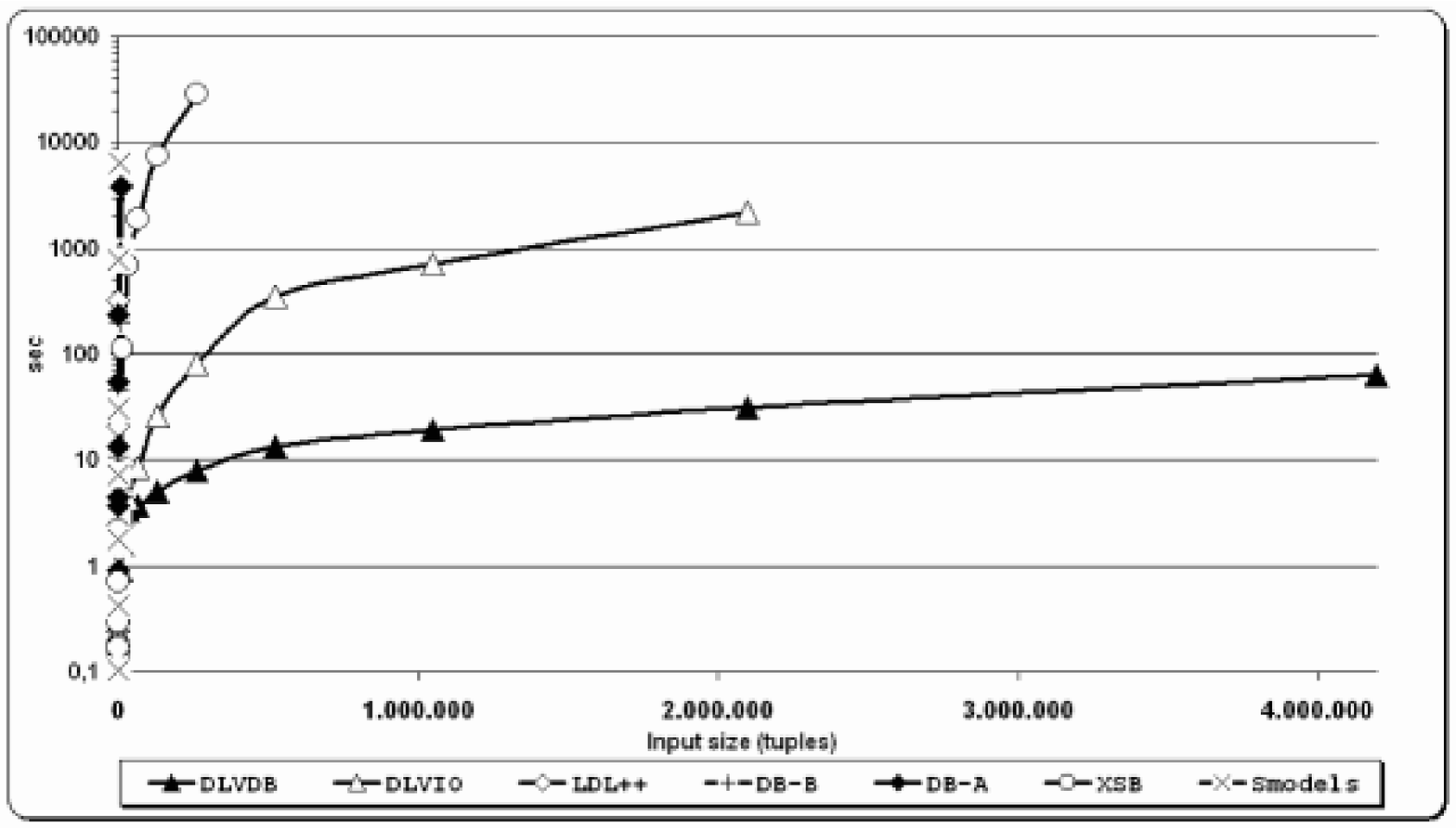,width=6.0cm,height=4.0cm} \\

\small{samegen(b1,b2)} & \small{reachable(X,Y) on a-graphs}\\
\psfig{figure=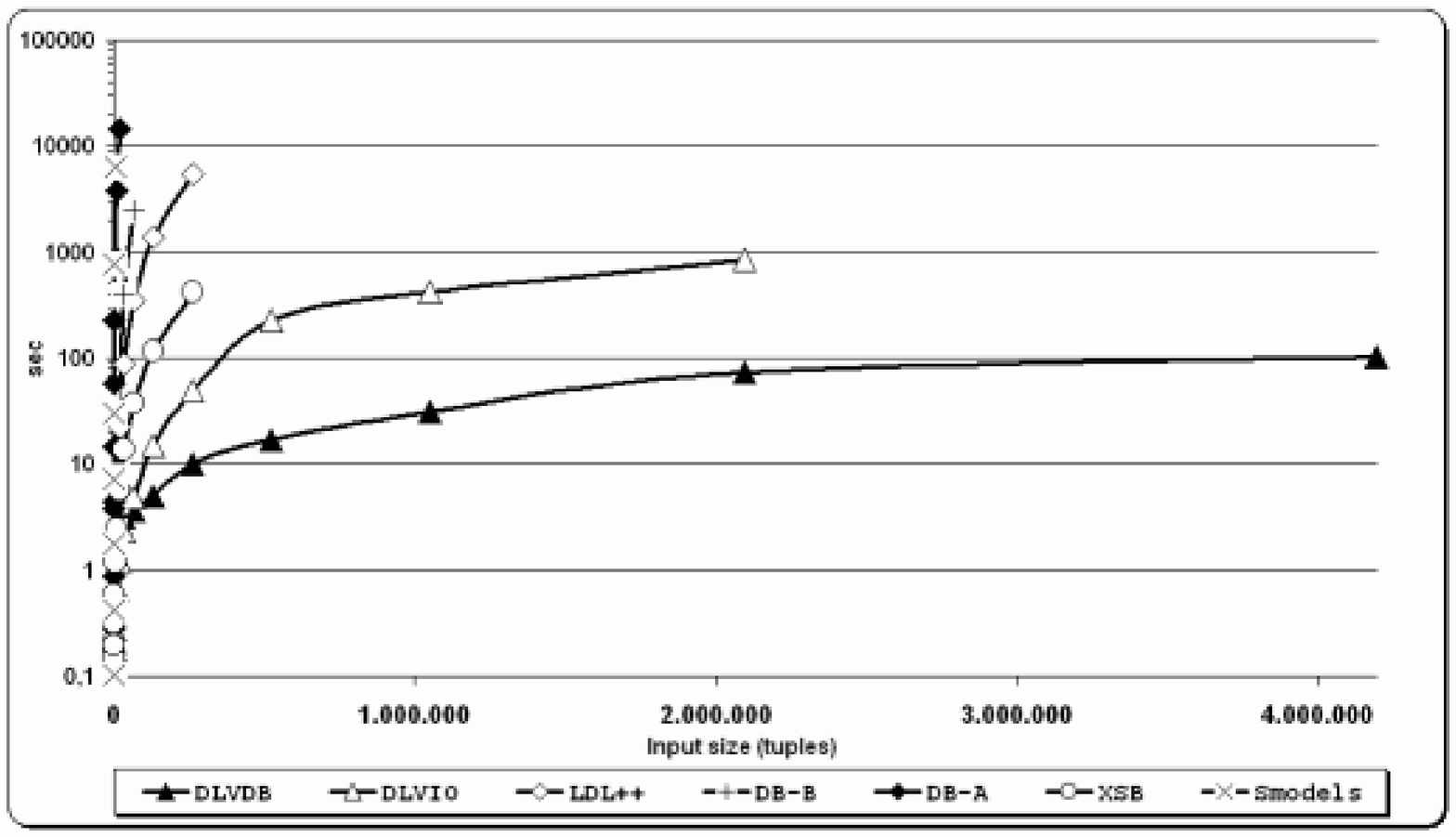,width=6.0cm,height=4.0cm} &
\psfig{figure=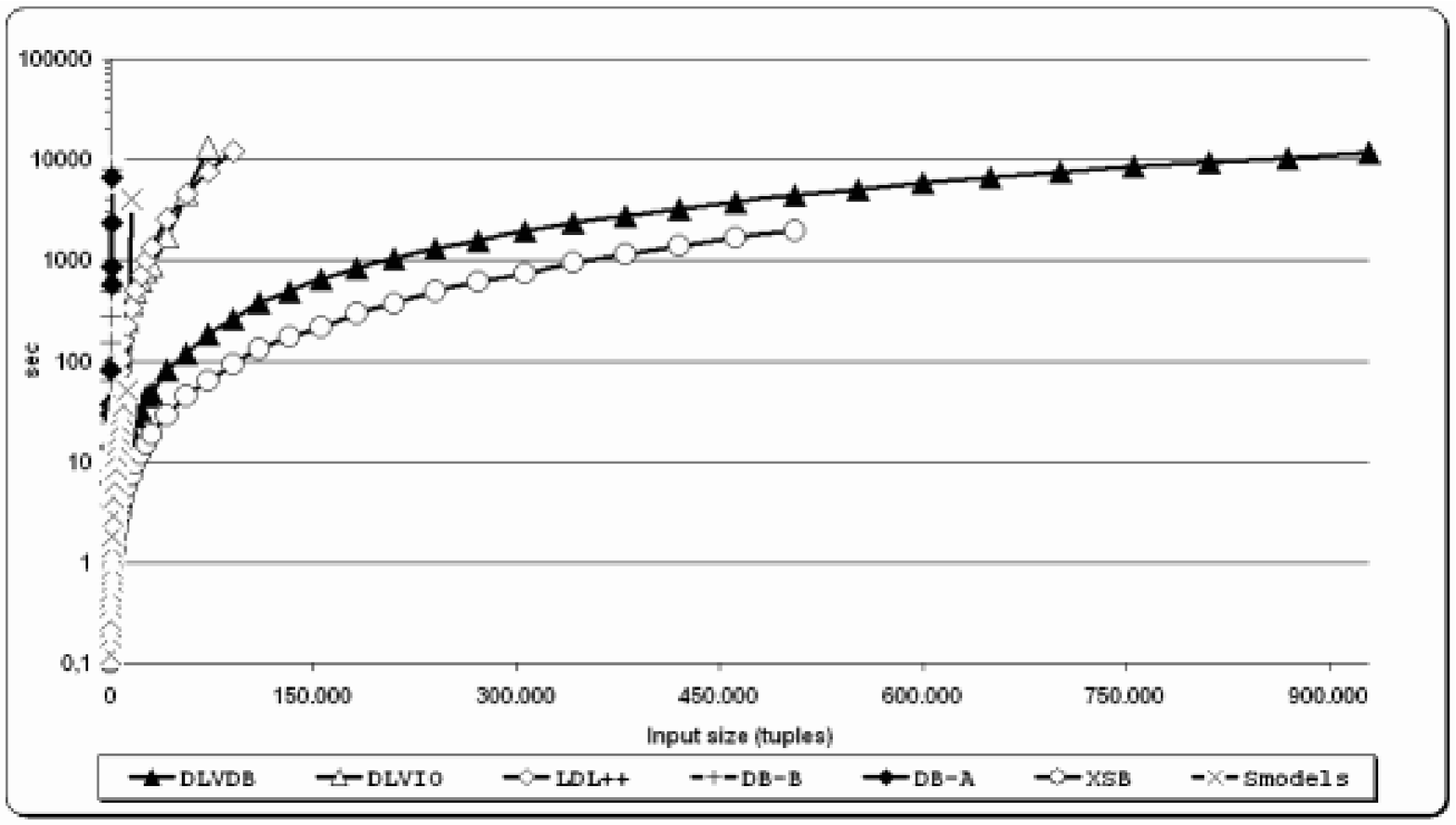,width=6.0cm,height=4.0cm}\\

\small{reachable(b1,Y) on a-graphs} & \small{reachable(b1,b2) on a-graphs}\\
\psfig{figure=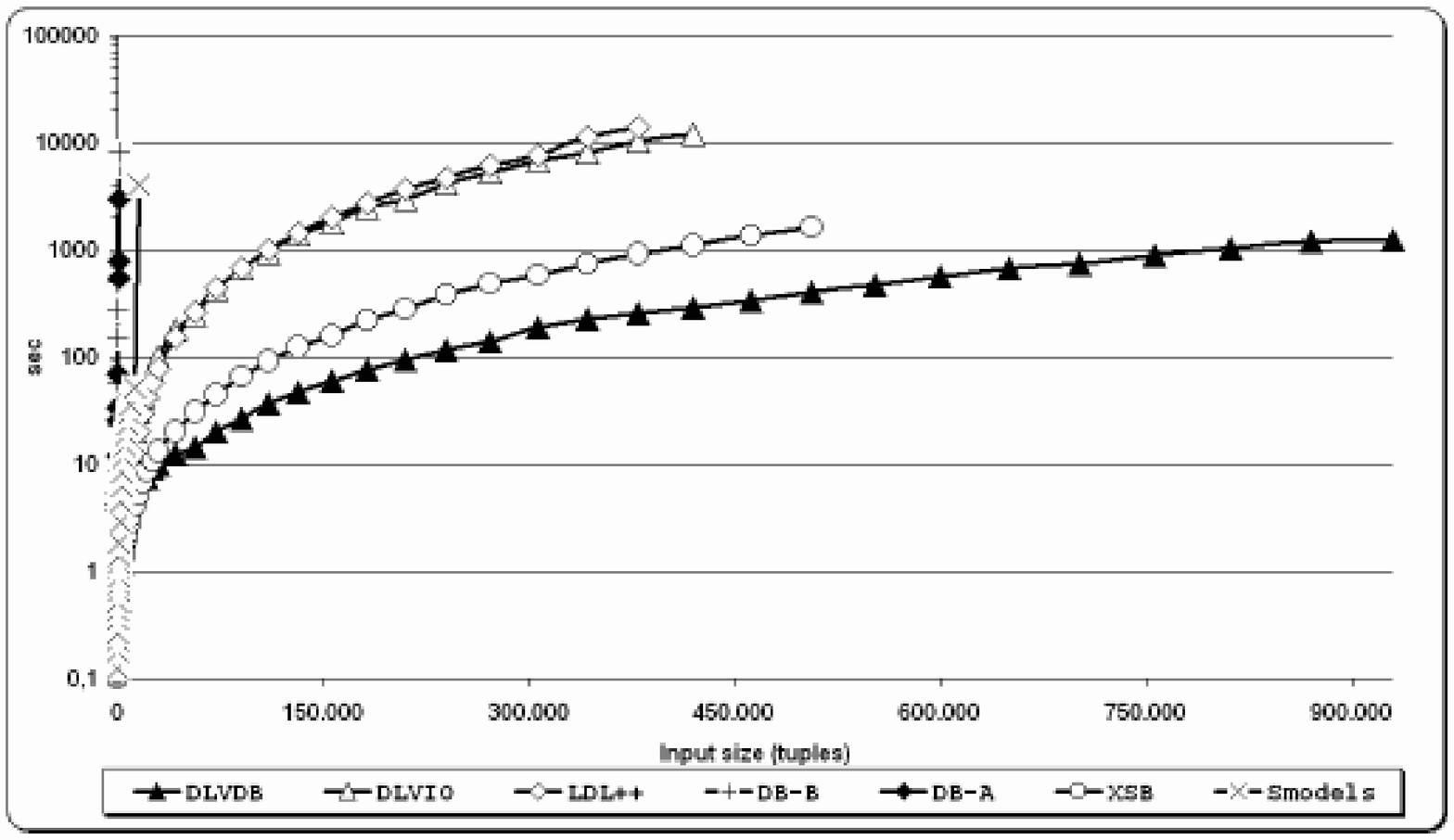,width=6.0cm,height=4.0cm} &
\psfig{figure=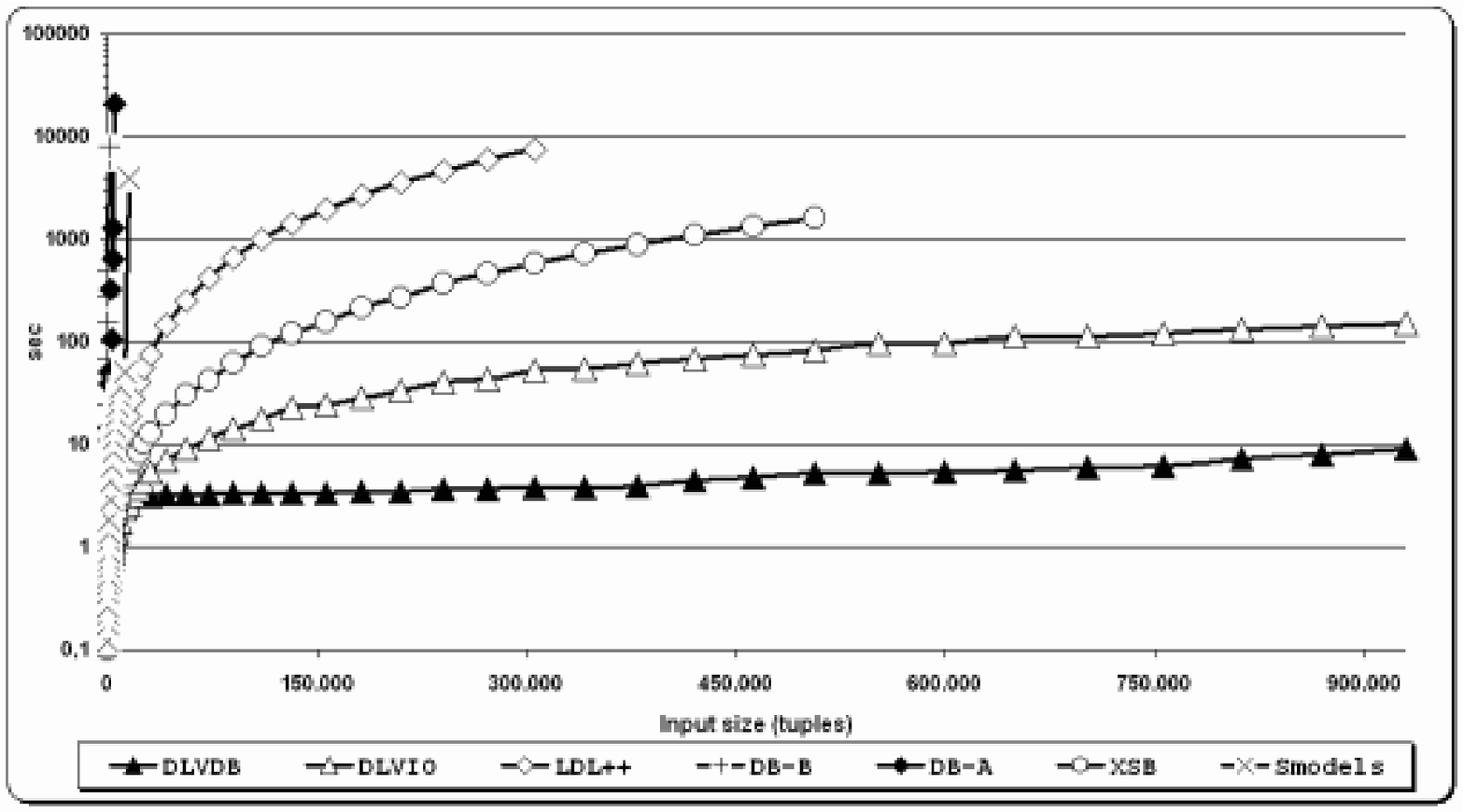,width=6.0cm,height=4.0cm} \\
\end{tabular}
\caption{Results for Same Generation on trees and Reachability with acyclic graphs}
\label{fig:samegen-reach-acyclic}
\end{figure}

\begin{figure}
\begin{tabular}{cc}
\small{reachable(X,Y) on c-graphs} & \small{reachable(b1,Y) on c-graphs} \\
\psfig{figure=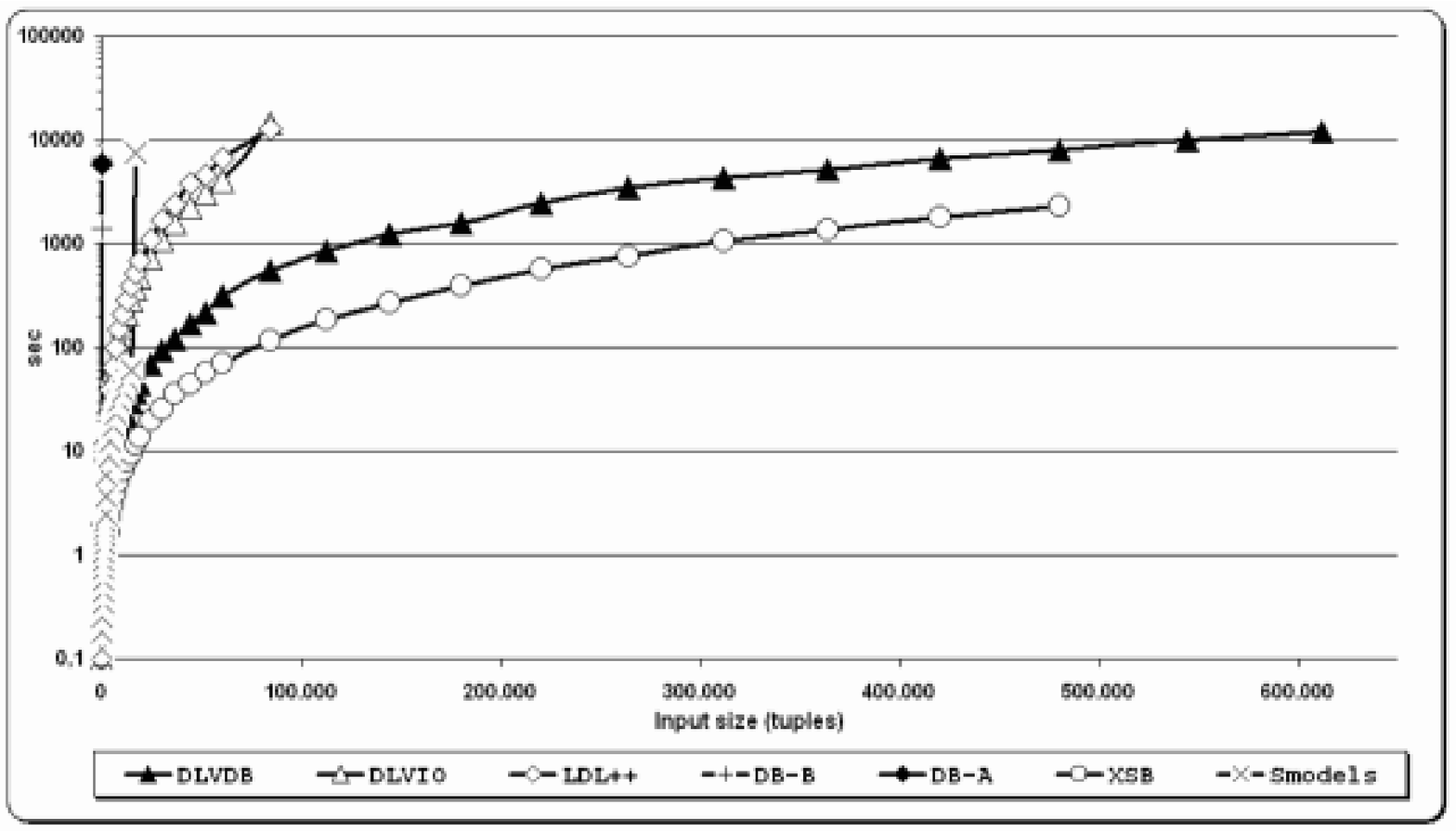,width=6.0cm,height=4.0cm} &
\psfig{figure=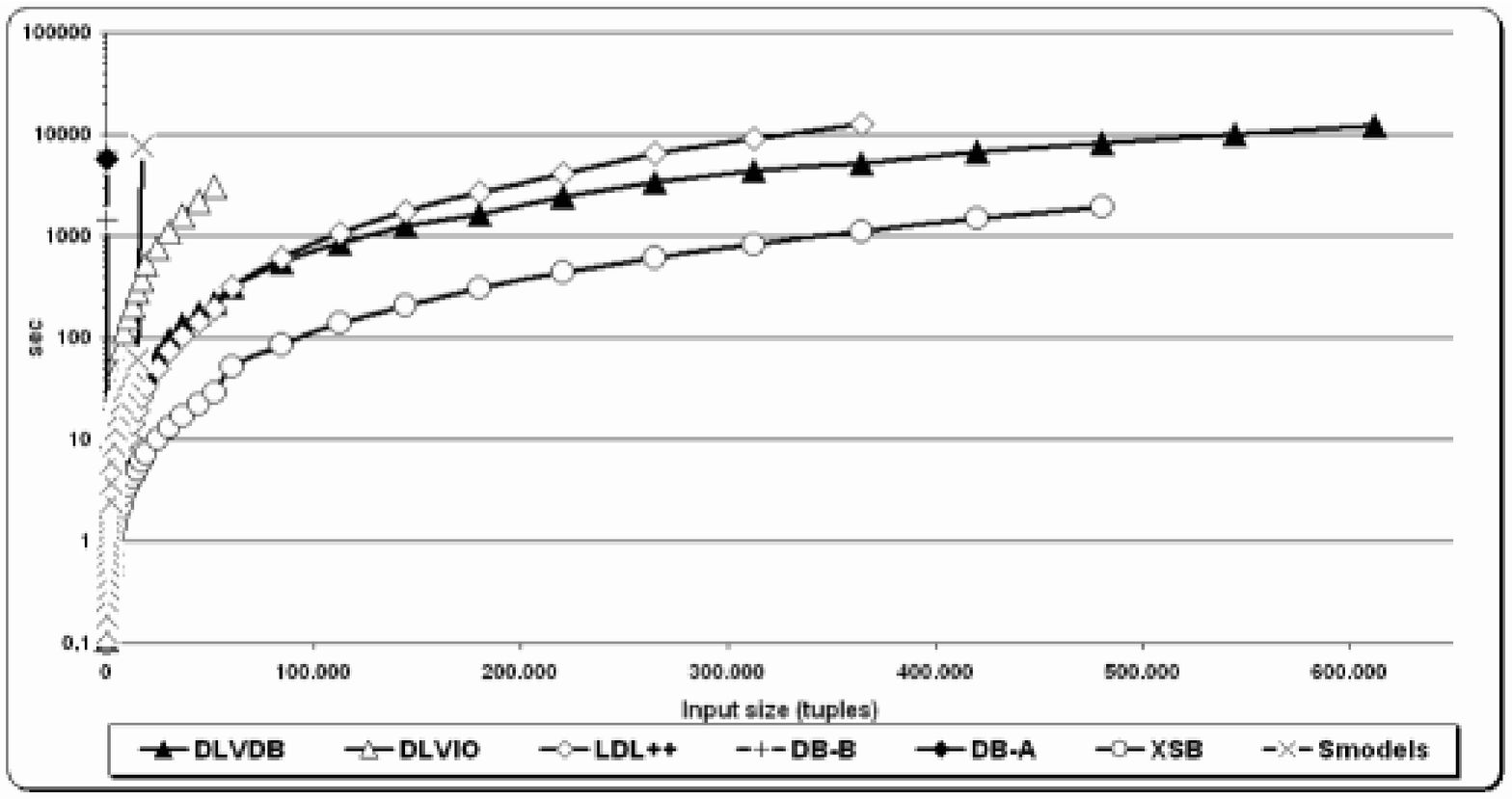,width=6.0cm,height=4.0cm} \\

\small{reachable(b1,b2) on c-graphs} & \small{reachable(X,Y) on cylinders} \\
\psfig{figure=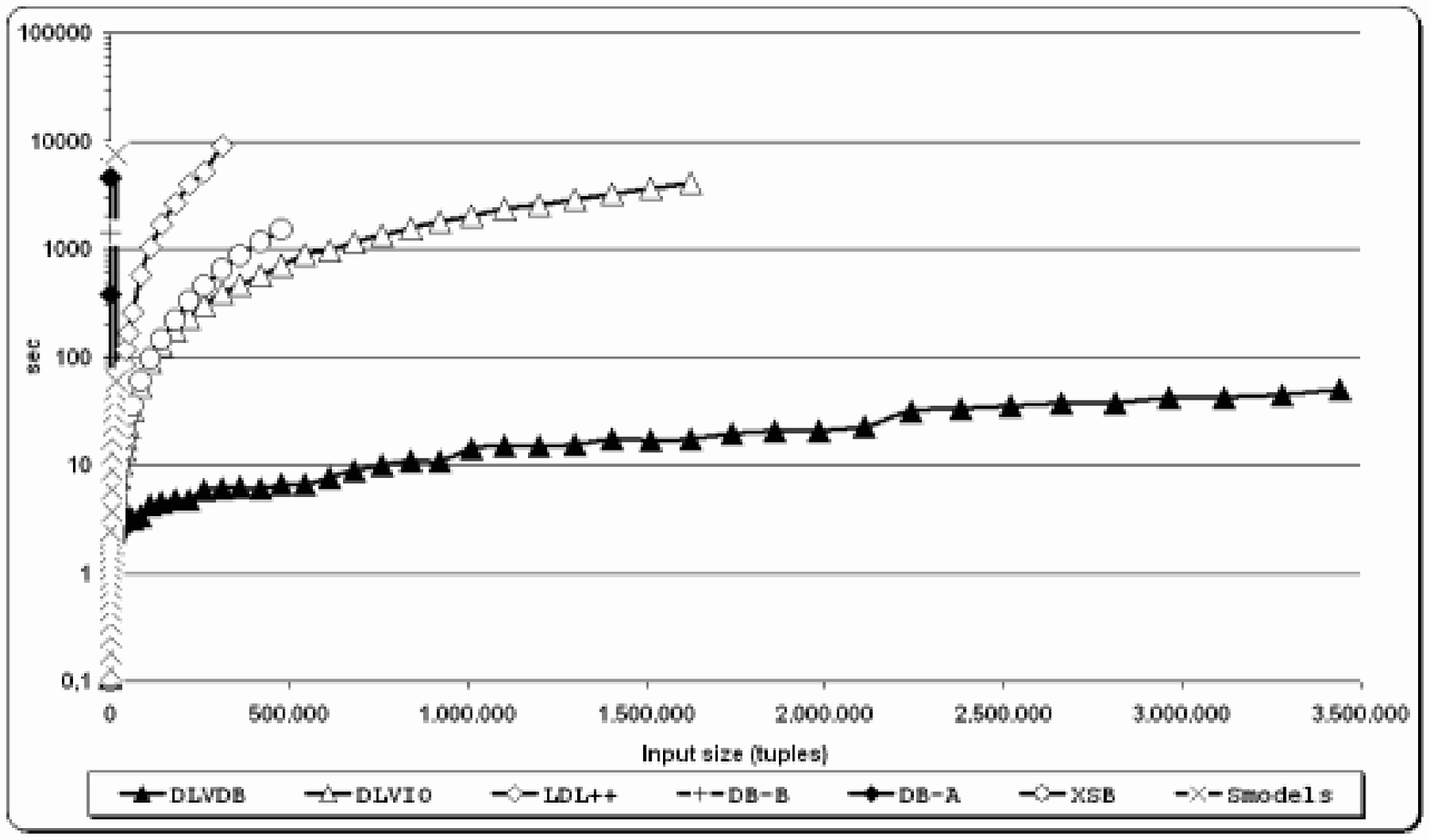,width=6.0cm,height=4.0cm}  &
\psfig{figure=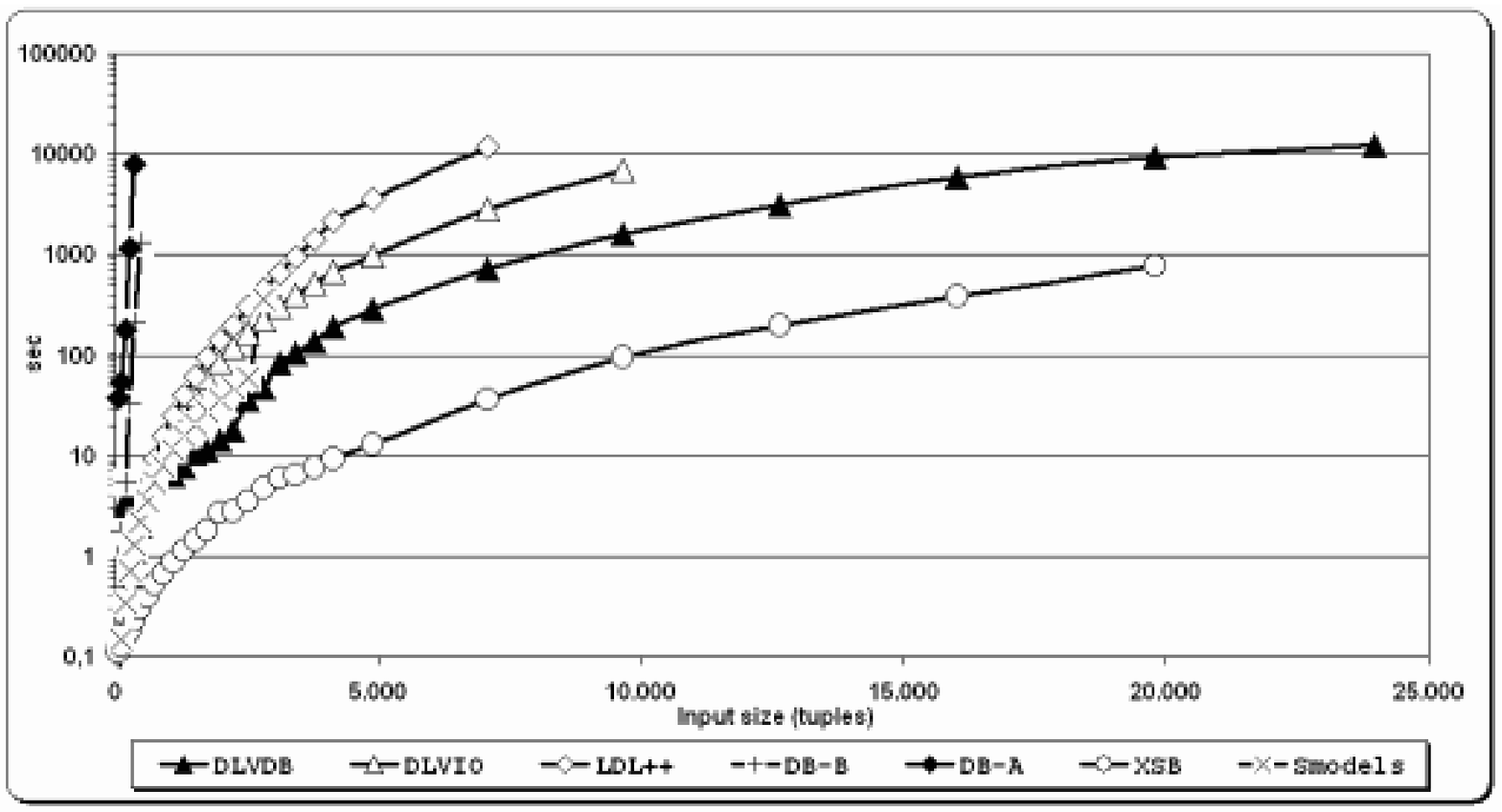,width=6.0cm,height=4.0cm} \\

\small{reachable(b1,Y) on cylinders}  & \small{reachable(b1,b2) on cylinders}\\
\psfig{figure=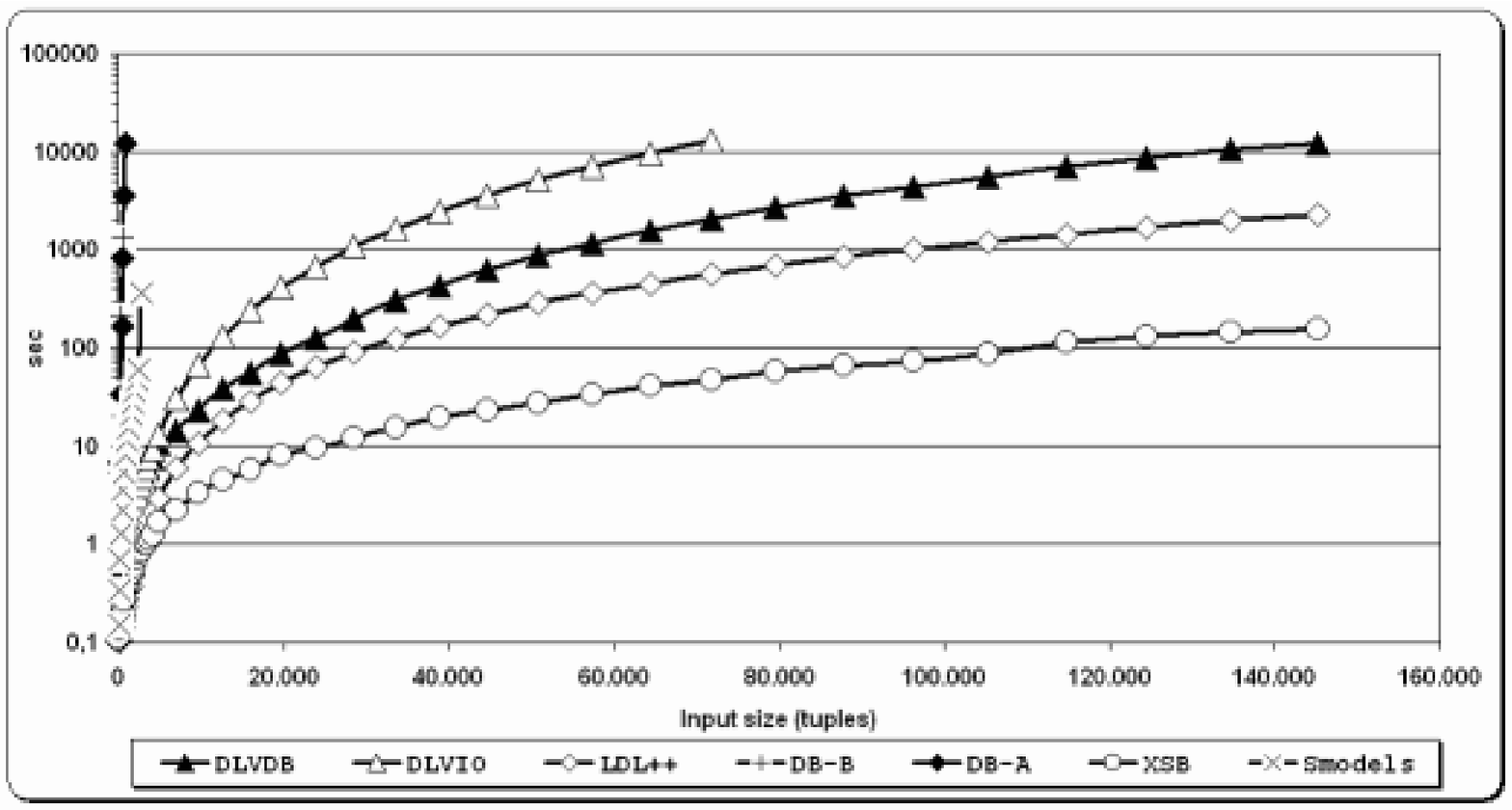,width=6.0cm,height=4.0cm} &
\psfig{figure=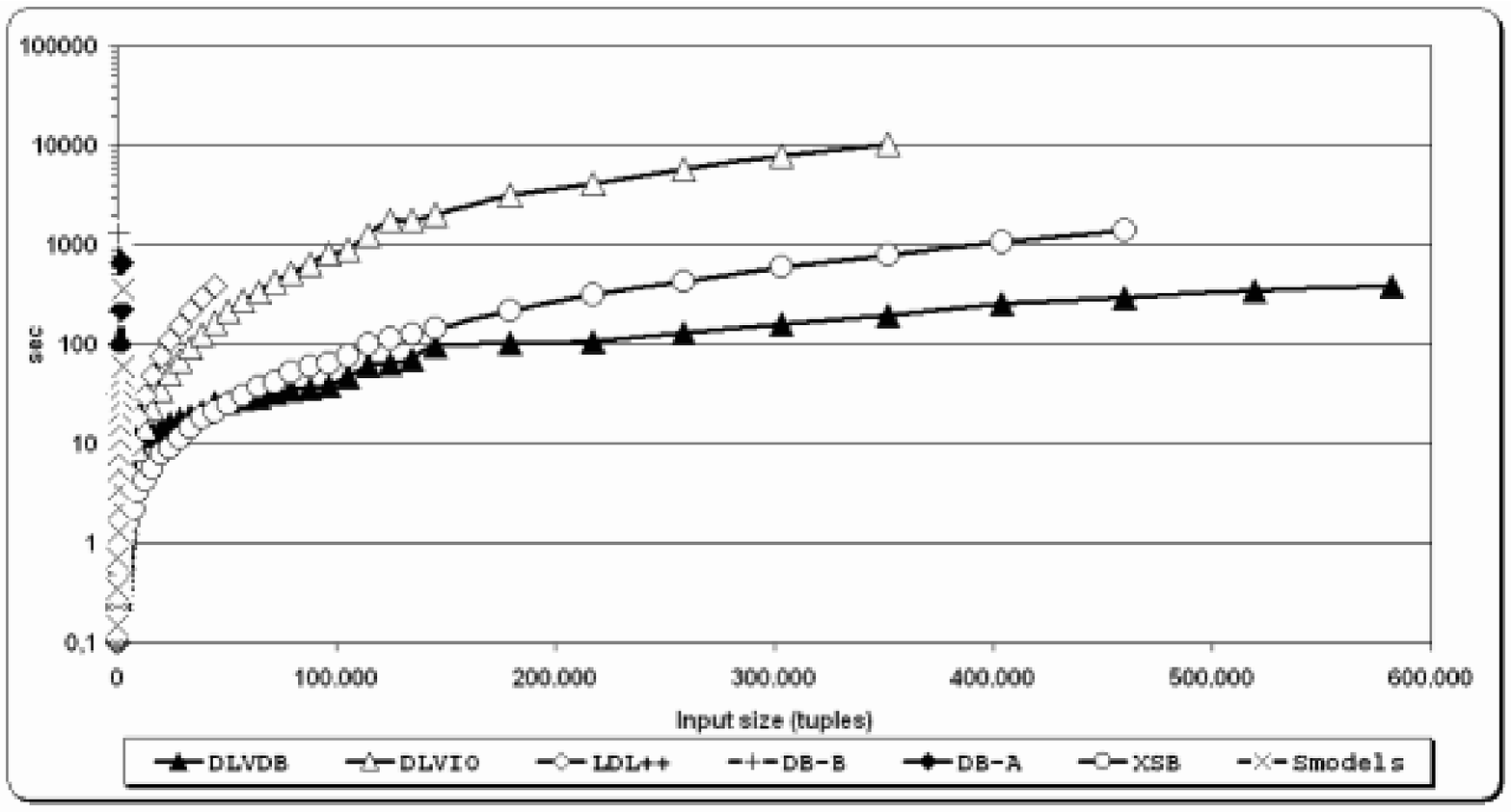,width=6.0cm,height=4.0cm} \\
\end{tabular}
\caption{Results for Reachability with cyclic graphs and with cylinders}
\label{fig:reach-c-graphs-cylinder}
\end{figure}

\begin{figure}
\begin{tabular}{cc}
\small{reachable(X,Y) on trees}  & \small{reachable(b1,Y) on trees}\\
\psfig{figure=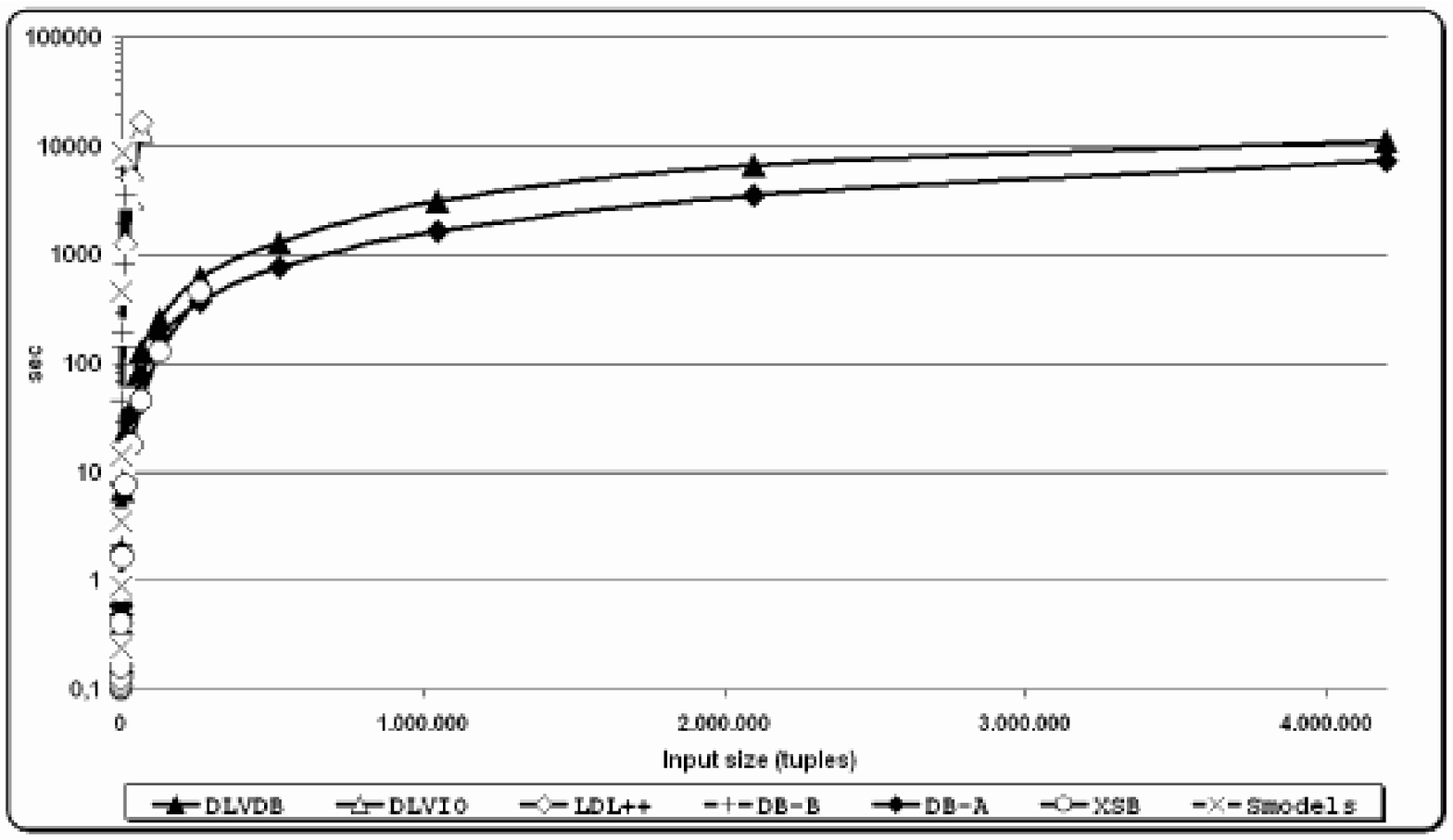,width=6.0cm,height=4.0cm} &
\psfig{figure=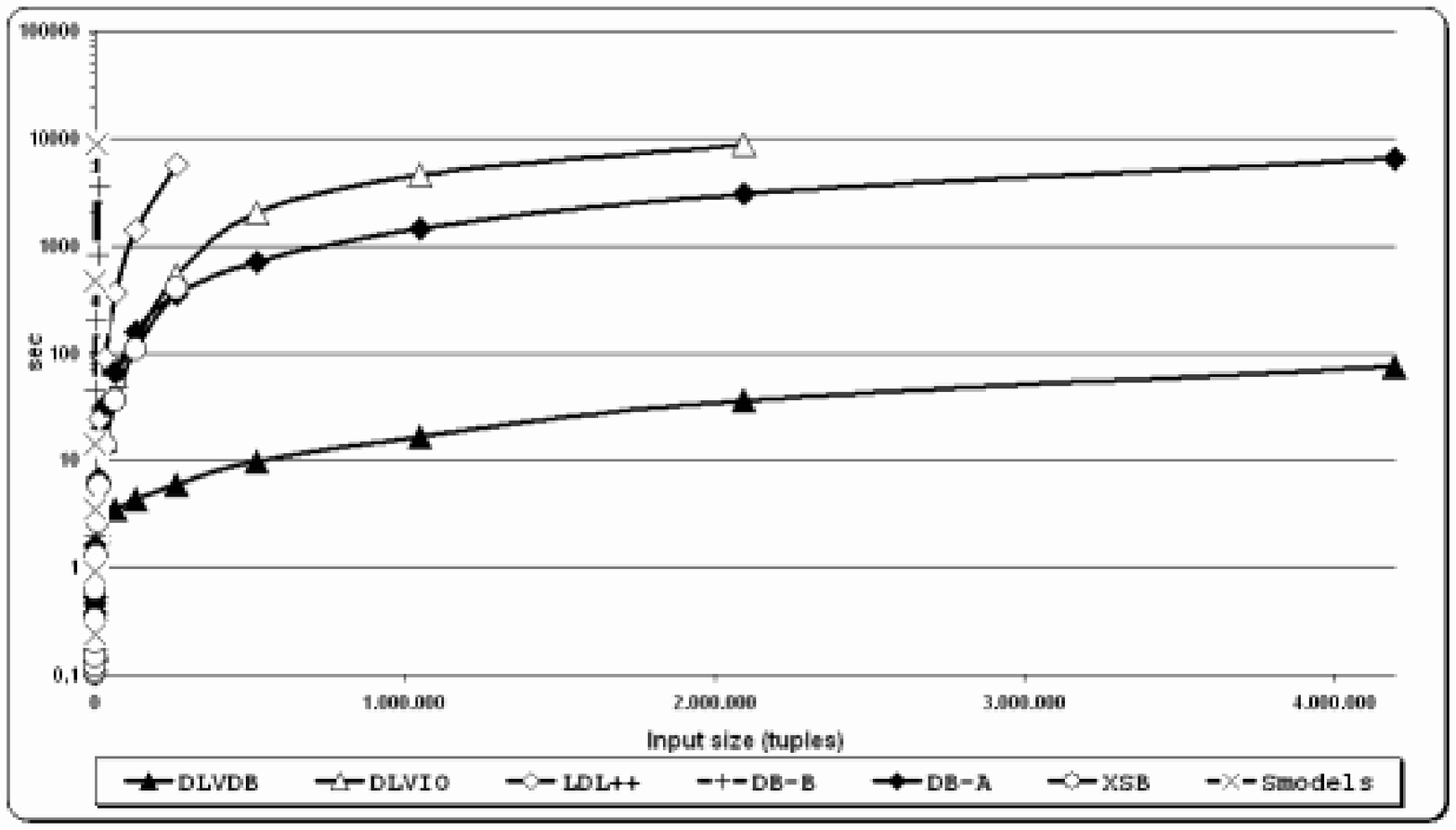,width=6.0cm,height=4.0cm} \\

\small{reachable(b1,b2) on trees} & \\
\psfig{figure=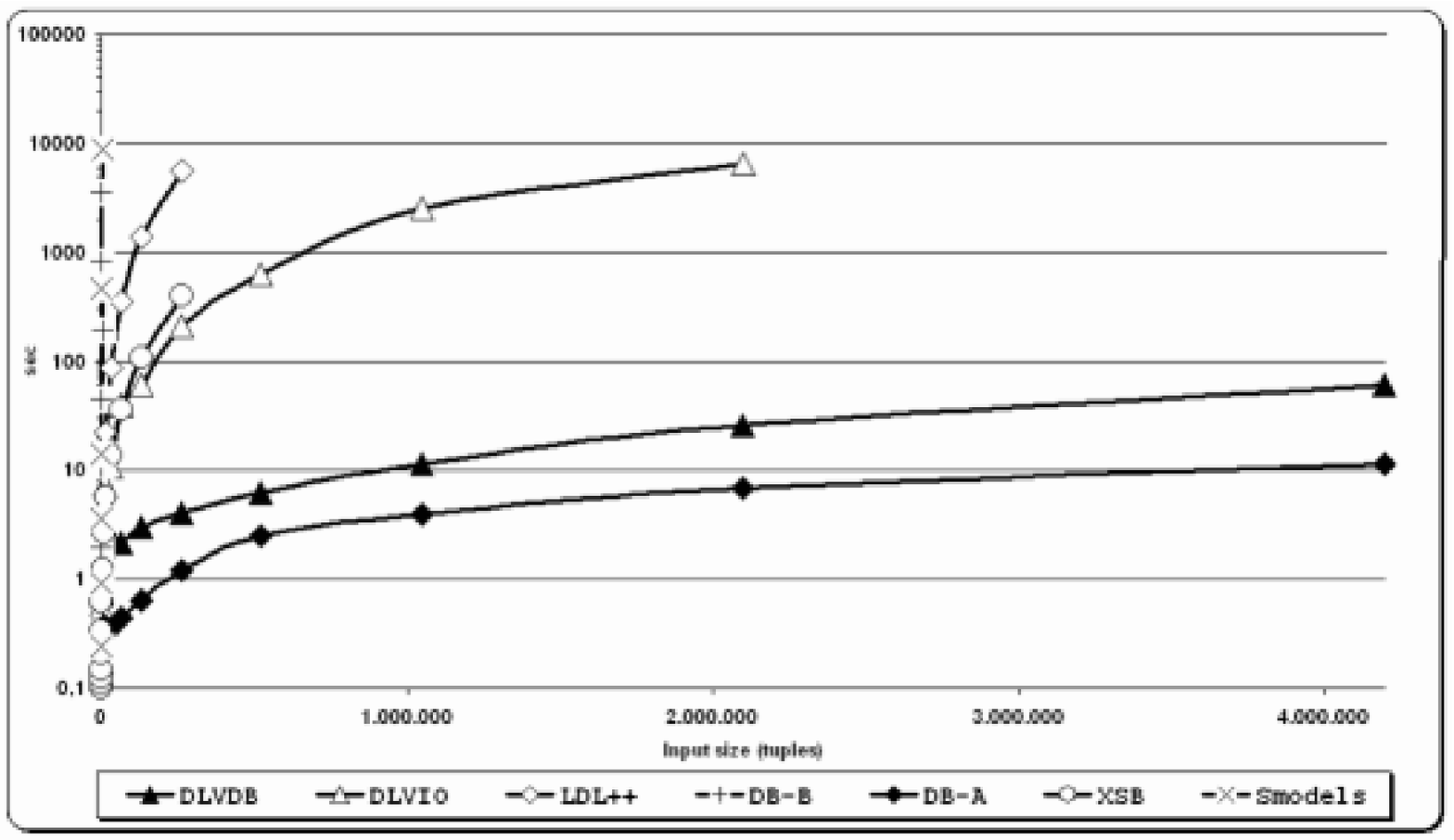,width=6.0cm,height=4.0cm} \\
\end{tabular}
\caption{Results for Reachability with trees} \label{fig:reach-tree}
\end{figure}

In more detail, Figures \ref{fig:samegen-reach-acyclic}-\ref{fig:reach-tree} show results
obtained for the various tests; the headline of each graph reports the corresponding query.

From the analysis of these figures we can observe that, in several cases, the performance of
\dlvdb (the black triangle in the graphs) is better than all the other systems with orders of
magnitude and that \dlvdb allows almost always the handling of the greatest amount of data;
moreover, there is no system which can be considered the ``competitor'' of \dlvdb in \emph{all}
the tests.

In particular, in some tests, XSB shows a good behaviour (e.g., in Reachability on cyclic graphs
and cylinders) but, even in those positive tests, it ``dies'' earlier than \dlvdb (with the
exception of reachable(b1,Y) on cylinders), probably because it exceeds the main-memory.

LDL++ is competitive with \dlvdb only in reachable(b1,Y) on cyclic graphs and cylinders, whereas
in all the other queries the performance difference is of more than one order of magnitude.

DB-B performance is near to that of \dlvdb only in samegen(X,Y); in all the other cases its line
is near to the vertical axis.

DB-A showed very good performance only for reachability on trees (see also Table
\ref{Tab:MaxSizes} introduced next). This behaviour could be justified by the presence of
optimization mechanisms implemented in this system which are particularly suited for computing
the transitive closure on simple data structures (like trees), but these are not effective for
other (more complex) kinds of query/data structure.

Surprisingly enough, DBMSs often have  the worst performance (their times are near to the
vertical axis) and they can handle very limited amounts of input data.

Finally, as expected, \dlvio is capable of handling lower amounts of data w.r.t. \dlvdb; however,
in several cases it was one of the best three performing systems, especially on bound queries.
This result is mainly due to the magic sets optimization technique it implements.

A rather surprising result is that \dlvio has almost always higher execution times than \dlvdb
even for not very high input data sizes. The motivation for this result can be justified by the
following reasoning. Both \dlvdb and \dlvio benefit from all the program rewriting optimization
techniques developed in the \dlv project; moreover, both of them implement a differential
Semi-Naive approach for the evaluation of normal stratified programs. However, while \dlvio
reasons about its underlying data in a tuple-at-a-time way, \dlvdb exploits a set-at-a-time
strategy (implemented by SQL queries); this, in conjunction with the fact that the underlying
working database implements advanced data-oriented optimization strategies, makes \dlvdb more
efficient than \dlvio even when all the data fits in main-memory.

As pointed out also in \cite{ban-ram-88}, another important parameter to measure in this context
is the system's capability of handling large amounts of data. In order to carry out this
verification, we considered the time response of each system for the largest input data set we
have used in each query.

Table \ref{Tab:MaxSizes} shows the execution times measured for those systems which have been
capable of solving the query within the fixed time limit of 12000 seconds; the second column of
the table shows, for each query, both the input data size, measured in terms of the number of
input facts (tuples), and the total amount of handled data, measured in Mbytes, given by the size
of the answer set produced by \dlvdb in answering that query\footnote{Note that all facts
produced by \dlvdb to answer the query are considered}.

From the analysis of this table, we may observe that: {\em (i)} \dlvdb has been always capable of
solving the query on the maximum data size; {\em (ii)} in 11 queries out of 15 \dlvdb (in one
case along with \dlvio) has been the only system capable of completing the computation within the
time limit; {\em (iii)} \dlvdb allowed to handle up to 6.7 Gbytes of data in samegen(X,Y) and 1.6
Gbytes in reachable(X,Y) on trees within the fixed time limit of 12000 seconds and never ended
its computation due to lack of memory, as instead other systems did.

\begin{table}[t]
\vspace*{-0.0cm}
 \caption{Execution times of the systems capable of solving the query for the
maximum considered size of the input data} \label{Tab:MaxSizes}
 \vspace*{-0.0cm}
{\tiny
\begin{center}

\begin{tabular}{lc||ccccccc}

\hline \hline

         Query /      &  \hspace*{-0.5cm}  Input Size (tuples)  /   & DB-B   & \hspace*{-0.1cm} DLV$^{IO}$ & \hspace*{-0.1cm} DLV$^{DB}$ & \hspace*{-0.1cm} LDL++ & \hspace*{-0.1cm} Smodels & \hspace*{-0.1cm} DB-A   & \hspace*{-0.1cm} XSB   \\

        {\em Data Type}   &   \hspace*{-0.5cm}{\em Output size} ({\em Mbytes}) & (sec) & \hspace*{-0.1cm} (sec)& \hspace*{-0.1cm} (sec)  & \hspace*{-0.1cm} (sec)  & \hspace*{-0.1cm} (sec)  & \hspace*{-0.1cm} (sec)   & \hspace*{-0.1cm} (sec) \\
\hline


           samegen(X,Y)     & 32766        & $-$   & $-$  & 5552  & $-$   & $-$     & $-$     & $-$   \\
           {\em tree} & {\em 6716 Mb}& & & & & & & \\
           samegen(b1,Y)    & 4194302       & $-$   & $-$   & 64      & $-$   & $-$     & $-$     & $-$   \\
           {\em tree} & {\em 78 Mb} & & & & & & & \\
           samegen(b1,b2)   & 4194302       & $-$   & $-$   & 102      & $-$ & $-$     & $-$     & $-$   \\
           {\em tree} & {\em 78 Mb} & & & & & & & \\
           reachable(X,Y)     & 929945       & $-$   & $-$  & 11820 & $-$   & $-$     & $-$     & $-$   \\
           {\em a-graph} & {\em 103 Mb} & & & & & & & \\
           reachable(b1,Y)    & 929945       & $-$   & $-$  & 1191  & $-$   & $-$     & $-$     & $-$   \\
           {\em a-graph}& {\em 38 Mb} & & & & & & & \\
           reachable(b1,b2)   & 929945       & $-$   & $-$  & 4      & $-$   & $-$     & $-$     & $-$   \\
           {\em a-graph} & {\em 17 Mb}& & & & & & & \\
           reachable(X,Y)    & 612150       & $-$   & $-$  & 11936  & $-$   & $-$     & $-$     & $-$   \\
           {\em c-graph} & {\em 68 Mb} & & & & & & & \\
           reachable(b1,Y)   & 612150       & $-$   & $-$  & 11933 & $-$   & $-$     & $-$     & $-$   \\
           {\em c-graph} & {\em 68 Mb} & & & & & & & \\
            reachable(b1,b2)     & 612150       & $-$   & 981  & 8      & $-$   & $-$     & $-$     & $-$   \\
            {\em c-graph} & {\em 11 Mb} & & & & & & & \\
           reachable(X,Y)          & 23980        & $-$   & $-$  & 11784  & $-$   & $-$     & $-$     & $-$   \\
           {\em cylinder} & {\em 465 Mb}& & & & & & & \\
           reachable(b1,Y)         & 145260       & $-$  & $-$  & 11654  & 2284   & $-$     & $-$     & 157   \\
           {\em cylinder} & {\em 279 Mb}& & & & & & & \\
           reachable(b1,b2)       & 582120        & $-$   & $-$  & 388  & $-$  & $-$     & $-$     & $-$   \\
           {\em cylinder} & {\em 13 Mb}& & & & & & & \\
           reachable(X,Y)   & 4194302     & $-$   & $-$  & 11161 & $-$   & $-$     & 7280   & $-$   \\
           {\em tree} & {\em 1634Mb} & & & & & & & \\
           reachable(b1,Y)   & 4194302     & $-$   & $-$  & 76     & $-$   & $-$     & 6438   & $-$   \\
           {\em tree} & {\em 79 Mb} & & & & & & & \\
           reachable(b1,b2)   & 4194302     & $-$   & $-$  & 60    & $-$   & $-$     & 12      & $-$   \\
           {\em tree} & {\em 78 Mb} & & & & & & & \\

 \hline \hline
\end{tabular}

 \vspace*{-0.0cm}
\vspace{-2\baselineskip}
\end{center}
}
\end{table}

\subsection{Comparison to DB-C}

As previously pointed out, DB-C does not support the standard SQL99 encoding for recursive
queries, but it exploits a proprietary language for implementing a simplified form of recursion.
This language is less expressive than SQL99 for recursion; as an example, unbound recursive
queries cannot be implemented in DB-C; analogously, it does not allow to write recursive views in
a ``uniform'' way (i.e., independently from the specific bound parameters).

As for the problems addressed in this paper, it was not possible to write the unbound query
either for Reachability, or for Same Generation with DB-C. The other queries have encodings not
equivalent to the general version we adopted for the other systems.

As an example, the query $\q_{1}=reachable(b1$,$Y)$ can be expressed in DB-C by the following
statement:

\begin{quote}
\small{
\hspace*{0.1cm}SELECT $b1$, $edge.att_{2}$ FROM edge\\
\hspace*{0.45cm}START WITH $att_{1}$= $b1$ CONNECT BY PRIOR $att_{2}$ = $att_{1}$ }
\end{quote}

\noindent which, however, is equivalent to the datalog program:

\begin{quote}
$ \rel{reached}(b1).$

$\rel{reached}(\attr{X})\ \derives \rel{reached}(\attr{Y}), \
\rel{edge}(\attr{Y},\attr{X}). $

$\rel{reachable}(\attr{b1},\attr{Y})\ \derives \ \rel{reached}(\attr{Y}). $
\end{quote}

This is clearly a program that can be evaluated more easily than the general encoding, because it
involves a recursive rule with one single attribute and a unique starting point for the recursion
(the fact \emph{reached(b1)}); however, this query (and the equivalent program) is less general
than the one introduced in Section \ref{sub:Encoding}, since its structure must be modified if,
for example, we need to have both the parameters bound or if we want to bound the second
parameter instead of the first.

Clearly, testing such encodings against the other, more general, ones would have been unfair.
Anyway, we carried out some tests involving DB-C, by applying its encodings and the corresponding
datalog programs on the maximum data instances we considered for the various queries, in order to
have a rough idea on the performance. As an example, for the query $\q_{1}=reachable(b1$,$Y)$
mentioned above, on a-graphs (resp., c-graphs) of size 929945 (resp., 612150) tuples  we have
measured that DB-C takes 22.5 (resp., 15.9) seconds, whereas \dlvdb takes 6.4 (resp., 5.6)
seconds. Analogously, for the query $\q_{1}=samegen(b1$,$Y)$, on trees of size 4194302 tuples,
DB-C requires 1329.4 seconds to terminate the computation, whereas \dlvdb requires 500.8 seconds.
DB-C performed better than \dlvdb only for Reachability on trees; also in this case, as we have
done for DB-A, we may conjecture that this behaviour is motivated by the particular optimization
techniques implemented in the system.

These results are representative of the overall performance we have measured for DB-C in our
benchmarks; on the one hand they confirm our claim that the encodings solvable by DB-C are very
different, also from a performance point of view, w.r.t. the general ones used in our benchmarks
(as an example, this is proved by the significantly lower timing measured for \dlvdb in
$reachable(b1$,$Y)$ w.r.t. the same query in the standard encoding); on the other hand, they
allow us to conclude that the same reasoning as that drawn in Section \ref{sub:results} about
\dlvdb performance is still valid.

\section{Conclusions} \label{sec:conclusions}
In this paper we have presented \dlvdb, a new deductive system for reasoning on massive amounts
of data. It presents features of an efficient DDS but also extends the capability of handling
data residing in external databases to a disjunctive logic programming system. A thorough
experimental validation showed that \dlvdb provides both important speed ups in the running time
of typical deductive queries and the capability to handle larger amounts of data w.r.t. existing
systems. Interestingly, the experimental results show that \dlvdb significantly outperforms both
commercial DBMSs and other  logic-based systems in the evaluation of recursive queries.

The key reason for the relevant performance improvement obtained by our system is the integration
of the following factors: {\em (i)} The idea to employ the efficient engine of a commercial DBMS
for rule evaluation, by translating logical rules in SQL statements (which are then executed by a
DBMS), allowing us to exploit the efficient data-oriented optimization techniques of relational
databases. {\em (ii)} The exploitation of advanced optimization techniques developed in the field
of deductive databases for logical query optimization (like, e.g., magic sets). {\em (iii)} A
proper combination and a well-engineered implementation of the above ideas. Moreover, the usage
of a purely mass-memory evaluation strategy, improves previous deductive systems eliminating, in
practice, any limitation in the dimension of the input data.

In the future we plan to extend the language supported by the direct database execution and to
exploit the system in interesting research fields, such as data integration and data warehousing.
Moreover, a mixed approach exploiting both \dlvdb and \dlvio executions to evaluate hard problems
partially on mass-memory and partially in main-memory will be explored.

$\ $\\
{\small \noindent {\bf Aknowledgments.} This work has been partially supported by the italian
''Ministero delle Attivit\`a Produttive'' under project ``Discovery Farm'' B01/0297/P 42749-13,
and by M.I.U.R. under project ``ONTO-DLV: Un ambiente basato sulla Programmazione Logica
Disgiuntiva per il trattamento di Ontologie'' 2521.}

\vspace*{-4mm}

 {\footnotesize
\newcommand{\SortNoOp}[1]{}

}

\end{document}